%% file: 00_main_jdssv_ver_for_arxiv.tex
\documentclass[article]{aaaa}

\usepackage{thumbpdf,lmodern}
\usepackage{bm}
\usepackage{todonotes} %% Remove this package when finishing the manuscript

\usepackage{framed}

\usepackage{booktabs}
\usepackage{amsmath}
\usepackage{subfig}

\definecolor{bgcolor}{rgb}{0.95,0.95,0.95}
\newcommand{\tr}{\mathrm{tr}}
\newcommand{\SuppURL}{\url{https://takanori-fujiwara.github.io/s/cnrl/}}

\def\equationautorefname~#1\null{Equation (#1)\null}

\usepackage{tikz}

\newcommand*\circled[1]{\tikz[baseline=(char.base)]{
            \node[shape=circle,draw,inner sep=0.7pt] (char) {\textsf{\small #1}};}}
\definecolor{shadecolor}{RGB}{235,235,235}
\newcommand{\bgcolored}[2]{\par\noindent\colorbox{shadecolor}{\parbox{#2}{#1}}}

\usepackage{multicol}
\DeclareMathOperator*{\argmax}{argmax}

\author{
Takanori Fujiwara\\University of California, Davis
\And
Jian Zhao\\University of Waterloo
\And
Francine Chen\\Toyota Research Institute
\AND
Yaoliang Yu\\University of Waterloo
\And
Kwan-Liu Ma\\University of California, Davis
}
\Plainauthor{Takanori Fujiwara, Jian Zhao, Francine Chen, Yaoliang Yu, Kwan-Liu Ma}

\title{Network Comparison with \\Interpretable Contrastive Network Representation Learning}
\Plaintitle{Network Comparison with Interpretable Contrastive Network Representation Learning}
\Shorttitle{Network Comparison with Interpretable cNRL}

\Abstract{
  \input{0_abstract}
}

\Plainkeywords{Contrastive learning, network representation learning, interpretability, network comparison}

\Address{
  Takanori Fujiwara\\
  Department of Computer Science\\
  University of California, Davis\\
  2136 Kemper Hall, One Shields Avenue, Davis, CA 95616, USA\\
  E-mail: \email{tfujiwara@ucdavis.edu}\\
  URL: \url{https://takanori-fujiwara.github.io/}
}

\begin{document}

%% - Use \proglang{}, \pkg{}, and \code{} markup throughout the manuscript.
%% - If such markup is in (sub)section titles, a plain text version has to be
%%   added as well.
%% - All software mentioned should be properly \cite-d.
%% - All abbreviations should be introduced.
%% - Unless the expansions of abbreviations are proper names (like "Journal of Statistical Software" above) they should be in sentence case (like "generalized linear models" below).

%% - In principle "as usual" again.
%% - When using equations (e.g., {equation}, {eqnarray}, {align}, etc.
%%   avoid empty lines before and after the equation (which would signal a new
%%   paragraph.
%% - When describing longer chunks of code that are _not_ meant for execution
%%   (e.g., a function synopsis or list of arguments), the environment {Code}
%%   is recommended. Alternatively, a plain {verbatim} can also be used.
%%   (For executed code see the next section.)

\input{00_notation_commands.tex}
\input{1_introduction.tex}
\input{2_problem_def.tex}

\input{3_analysis_example.tex}
\input{4_short_methodology.tex}
\input{7_related_work.tex}

\input{5_1_case_studies.tex}
\input{5_2_design_comparison.tex}
\input{8_conclusions.tex}

\section*{Acknowledgments}
This research was partially carried out at FXPAL.
This research is sponsored in part by the U.S. National Science Foundation through grants IIS-1741536 and IIS-1528203, and FXPAL through its internship program.

% All acknowledgments should be collected in this
% unnumbered section before the references. It may contain the usual information
% about funding and feedback from colleagues/reviewers/etc. Furthermore,
% information such as relative contributions of the authors may be added here
% (if any).

%% -- Bibliography -------------------------------------------------------------
%% - References need to be provided in a .bib BibTeX database.
%% - All references should be made with \cite, \citet, \citep, \citealp etc.
%%   (and never hard-coded). See the FAQ for details.
%% - JSS-specific markup (\proglang, \pkg, \code) should be used in the .bib.
%% - Titles in the .bib should be in title case.
%% - DOIs should be included where available.

% \cite, \citet, \citep, \citealp
% \cite, \citet: Name et al. (year)
% \citealp: Name et al. year
% \citep: (Name et al. year)

\bibliography{00_reference}

%% -- Appendix (if any) --------------------------------------------------------
%% - After the bibliography with page break.
%% - With proper section titles and _not_ just "Appendix".

\newpage

\begin{appendix}
\input{A_appendix}
\end{appendix}

\newpage
%% -----------------------------------------------------------------------------

\end{document}

%% file: 0_abstract.tex
Identifying unique characteristics in a network through comparison with another network is an essential network analysis task. For example, with networks of protein interactions obtained from normal and cancer tissues, we can discover unique types of interactions in cancer tissues. This analysis task could be greatly assisted by contrastive learning, which is an emerging analysis approach to discover salient patterns in one dataset relative to another. However, existing contrastive learning methods cannot be directly applied to networks as they are designed only for high-dimensional data analysis. To address this problem, we introduce a new analysis approach called \emph{contrastive network representation learning} (cNRL). By integrating two machine learning schemes, network representation learning and contrastive learning, cNRL enables embedding of network nodes into a low-dimensional representation that reveals the uniqueness of one network compared to another. Within this approach, we also design a method, named i-cNRL, which offers interpretability in the learned results, allowing for understanding which specific patterns are only found in one network. We demonstrate the effectiveness of i-cNRL for network comparison with multiple network models and real-world datasets. Furthermore, we compare i-cNRL and other potential cNRL algorithm designs through quantitative and qualitative evaluations. 

%% file: 00_notation_commands.tex
\newcommand{\Graph}[1]{G_#1}
\newcommand{\GraphT}{\Graph{T}}
\newcommand{\GraphB}{\Graph{B}}
\newcommand{\Adj}[1]{\mathbf{A}_#1}
\newcommand{\AdjT}{\Adj{T}}
\newcommand{\AdjB}{\Adj{B}}
\newcommand{\Attr}[1]{\mathbf{P}_#1}
\newcommand{\AttrT}{\Attr{T}}
\newcommand{\AttrB}{\Attr{B}}
\newcommand{\nNodes}[1]{n_#1}
\newcommand{\nNodesT}{\nNodes{T}}
\newcommand{\nNodesB}{\nNodes{B}}
\newcommand{\nAttrs}[1]{m_#1}
\newcommand{\nAttrsT}{\nAttrs{T}}
\newcommand{\nAttrsB}{\nAttrs{B}}
\newcommand{\nEdges}[1]{l_#1}
\newcommand{\nEdgesT}{\nEdges{T}}
\newcommand{\nEdgesB}{\nEdges{B}}
\newcommand{\nNRLFeats}{d}
\newcommand{\nCNRLFeats}{{d'}}
\newcommand{\FeatMat}[1]{\mathbf{X}_#1}
\newcommand{\FeatMatT}{\FeatMat{T}}
\newcommand{\FeatMatB}{\FeatMat{B}}
\newcommand{\ProjMat}{\mathbf{W}}
\newcommand{\ContRepr}[1]{\mathbf{Y}_#1}
\newcommand{\ContReprT}{\ContRepr{T}}
\newcommand{\ContReprB}{\ContRepr{B}}

\newcommand{\BaseFeat}{{\rm \bf x}}
\newcommand{\RelFunc}{f}
\newcommand{\RelFeatOpe}[2]{\Phi^{#1}_{#2}}
\newcommand{\F}{\mathcal{F}}
\newcommand{\DepthRelFunc}{h}
\newcommand{\SummaryMeasure}{S}
\newcommand{\Mean}{\rm mean}
\newcommand{\Sum}{\rm sum}
\newcommand{\Max}{\rm max}
\newcommand{\Lpnorm}{L^2{\rm norm}}
\newcommand{\Cov}[1]{\mathbf{C}_#1}
\newcommand{\CovT}{\Cov{T}}
\newcommand{\CovB}{\Cov{B}}
\newcommand{\ContParam}{\alpha}
\newcommand{\ContConst}{\epsilon}

\newcommand{\DolphinNwID}{N1}
\newcommand{\KarateNwID}{N2}
\newcommand{\RandomNwID}{N3}
\newcommand{\PriceNwID}{N4}
\newcommand{\PtoPNwID}{N5}
\newcommand{\PriceTwoNwID}{N6}
\newcommand{\EPriceNwID}{N7}
\newcommand{\CombinedAPMSNwID}{N8}
\newcommand{\LCMultiNwID}{N9}
\newcommand{\SchFirstDayNwID}{N10}
\newcommand{\SchSecondDayNwID}{N11}

\newcommand{\DolphinVsKarate}{F1}
\newcommand{\PriceVsRandom}{F2}
\newcommand{\RandomVsPrice}{F3}
\newcommand{\PtoPVsPriceTwo}{F4}
\newcommand{\PtoPVsEPrice}{F5}
\newcommand{\LCMultiVsCombinedAPMS}{F6}
\newcommand{\SchSecondVsFirst}{F7}

%% file: 1_introduction.tex
\section{Introduction}

Networks are commonly used to model various types of relationships in real-world applications, such as social networks~\citep{crnovrsanin2014visualization}, cellular networks~\citep{chen2004content}, and communication networks~\citep{bhanot2005optimizing}. 
Comparative analysis of networks is an essential task in practice, where we want to identify differentiating factors between two networks or the uniqueness of one network compared to another~\citep{emmert2016fifty,tantardini2019comparing}.
For instance, when a neuroscientist is studying the effect of Alzheimer's disease on a human brain~\citep{gaiteri2016genetic}, they want to compare the brain network of a patient with Alzheimer's disease to that of a healthy subject. 
Also, for collaboration networks of researchers in different fields~\citep{lariviere2006canadian}, an analyst in a funding agency may want to discover any unique ways of collaborations in the fields for decision making. 

Several approaches have been proposed for network comparison~\citep{tantardini2019comparing}. 
When two networks have the same node-set and the pairwise correspondence between nodes is known, we can compute a similarity between two networks (e.g., the Euclidean distance between two adjacency matrices). 
When the node-correspondence is unknown or does not exist, a network-statistics based approach is commonly used (e.g., the clustering coefficient, network diameter, or node degree distribution). 
Another popular approach is using graphlets~\citep{prvzulj2007biological}---small, connected, and non-isomorphic subgraph patterns in a graph (e.g., the complete graph of three nodes). 
The similarities of two networks can be characterized by comparing the frequency of appearance of each graphlet in each network~\citep{kwon2017would}. 

While the existing approaches can provide a (dis)similarity between networks, they compare networks only based on one selected measure (e.g., node degree), which is often insufficient.
Also, these approaches only provide network-level similarities, and thus cannot compare networks in more detailed levels (e.g., a node-level). 
Without such a detailed-level comparison, it is difficult to find which part of a network relates to its uniqueness. 

To address these challenges, we introduce a new approach that integrates the concept of contrastive learning~\citep{zou2013contrastive,abid2018exploring} together with network representation learning (NRL), which we call \emph{cNRL}.
Within cNRL, NRL enables the characterization of networks with comprehensive measures without overwhelming a user with information by embedding nodes into a low-dimensional space; contrastive learning allows for discovering unique patterns in one dataset relative to another\footnote{There are two  different machine learning schemes both called contrastive learning. 
One aims to learn an embedding from similar and dissimilar pairs of input samples so that similar samples are placed close together in the embedding space, and vice versa~\citep{crl,technologies9010002}. Contrastive learning in our work refers to the other scheme introduced by \citet{zou2013contrastive}, where the learning purpose is finding unique/salient factors in one group of samples compared to another group.}~\citep{abid2018exploring}. 
By leveraging the benefits of both, we can reveal unique patterns in one network by contrasting with another, in a thorough (i.e., using multiple essential measures to capture the network characteristics) and detailed (i.e., analyzing a node or subnetwork level) manner. 

With our approach, we consider the generality and interpretability of cNRL, and introduce a method called \emph{i-cNRL}. 
First, i-cNRL is designed not to require node-correspondences or network alignment~\citep{emmert2016fifty}, and thus is applicable to various networks.
Also, unlike many other NRL methods, such as node2vec \citep{grover2016node2vec} and graph neural networks~\citep{zhang2018deep}, i-cNRL offers interpretability~\citep{adadi2018peeking}, providing information about the meaning of an identified pattern and the reason why that pattern can be seen in only that network.

In summary, our main contributions include:
\begin{itemize}
    \item A new approach, called contrastive network representation learning (cNRL), which aims to reveal unique patterns in one network relative to another network.
    \item A method exemplifying cNRL, called i-cNRL, which (1) offers general applicability, including networks without node-correspondence or network alignment, (2) provides interpretability for helping understand revealed patterns, and (3) equips automatic hyperparameter selection for contrastive learning.
    \item Experiments using multiple network models and real-world datasets, which demonstrate the capability of comparative network analysis.
    \item Quantitative and qualitative comparisons with other potential designs of cNRL methods.
\end{itemize}
We provide \proglang{Python} implementations of cNRL and i-cNRL, datasets, and source code used for the evaluations at \SuppURL.

%% file: 2_problem_def.tex
\section{Problem Definition}
\label{sec:problem_def}

We here define the problem to be addressed by cNRL. 
Given two networks, a target network $\GraphT$ and a background network $\GraphB$, we want to seek unique patterns in $\GraphT$ relative to $\GraphB$. 
Similar to contrastive learning~\citep{zou2013contrastive}, the unique patterns can be represented as relationships (e.g., the structural differences among network nodes) that appear in $\GraphT$ but do not appear in $\GraphB$.

For example, when finding unique patterns in a scale-free network $\GraphT$ (i.e., its node-degree distribution follows a power law) relative to a random network $\GraphB$ (i.e., each node pair is connected with a fixed probability)~\citep{barabasi2016network}, we should be able to capture the unique patterns related to node degrees since $\GraphT$ has more variety in node degrees. 
For practical usage, the unique patterns could relate to more complicated centralities, measures, combinations of them, and many more.

Note that, as with the existing work of contrastive learning~\citep{zou2013contrastive,abid2018exploring}, cNRL does not aim to discriminate graph elements (i.e., nodes or links) in $\GraphT$ from $\GraphB$, but to identify unique patterns in $\GraphT$.
For example, when comparing scale-free and random networks, the node degree should be able to highlight the aforementioned unique patterns; however, it cannot well distinguish nodes in $\GraphT$ from ones in $\GraphB$ as many nodes in $\GraphT$ and $\GraphB$ could have similar degrees.

%% file: 3_analysis_example.tex
\section{Analysis Example}
\label{sec:analysis_example}

\begin{figure}[tb]
	\centering
	\captionsetup{farskip=0pt}
	\hspace*{-10pt}
    \subfloat[$\GraphT$: Dolphin]{
        \includegraphics[width=0.23\linewidth]{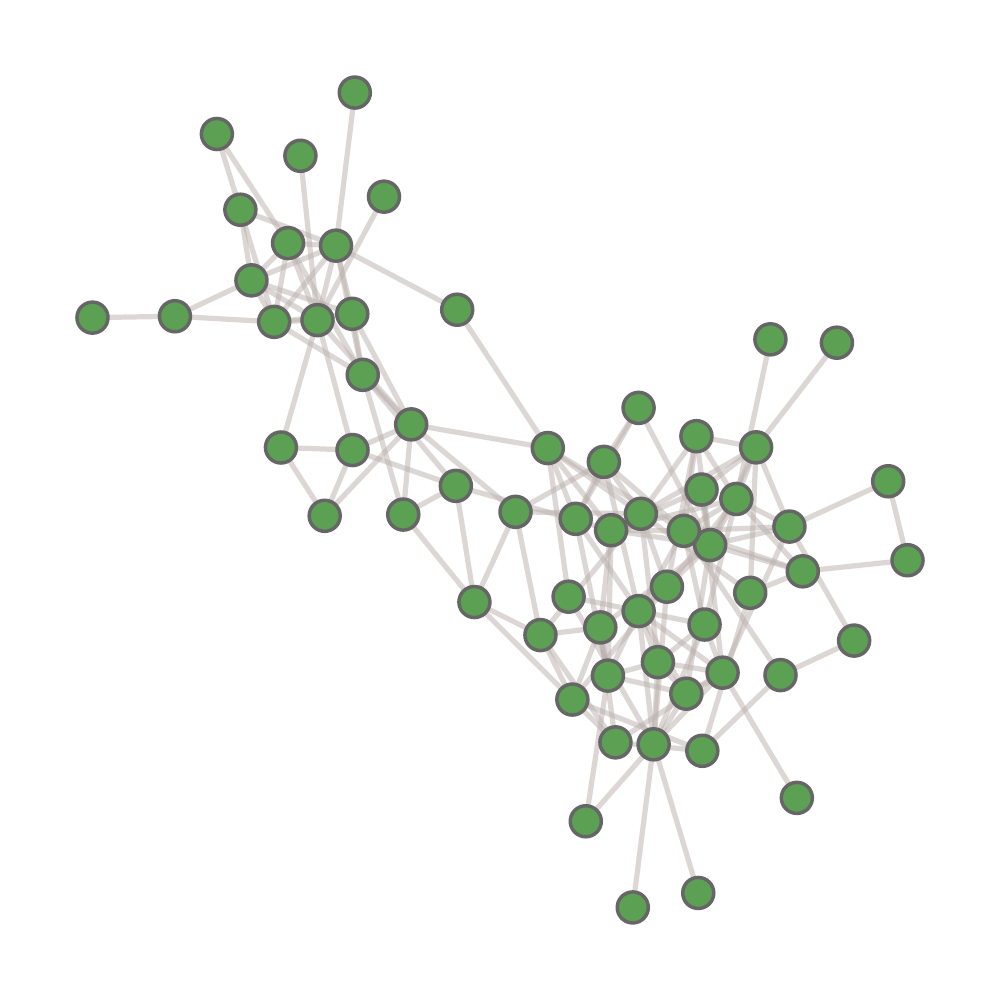}
        \label{fig:dolphin}
    }
    \hspace*{-10pt}
    \subfloat[$\GraphB$: Karate]{
        \includegraphics[width=0.23\linewidth]{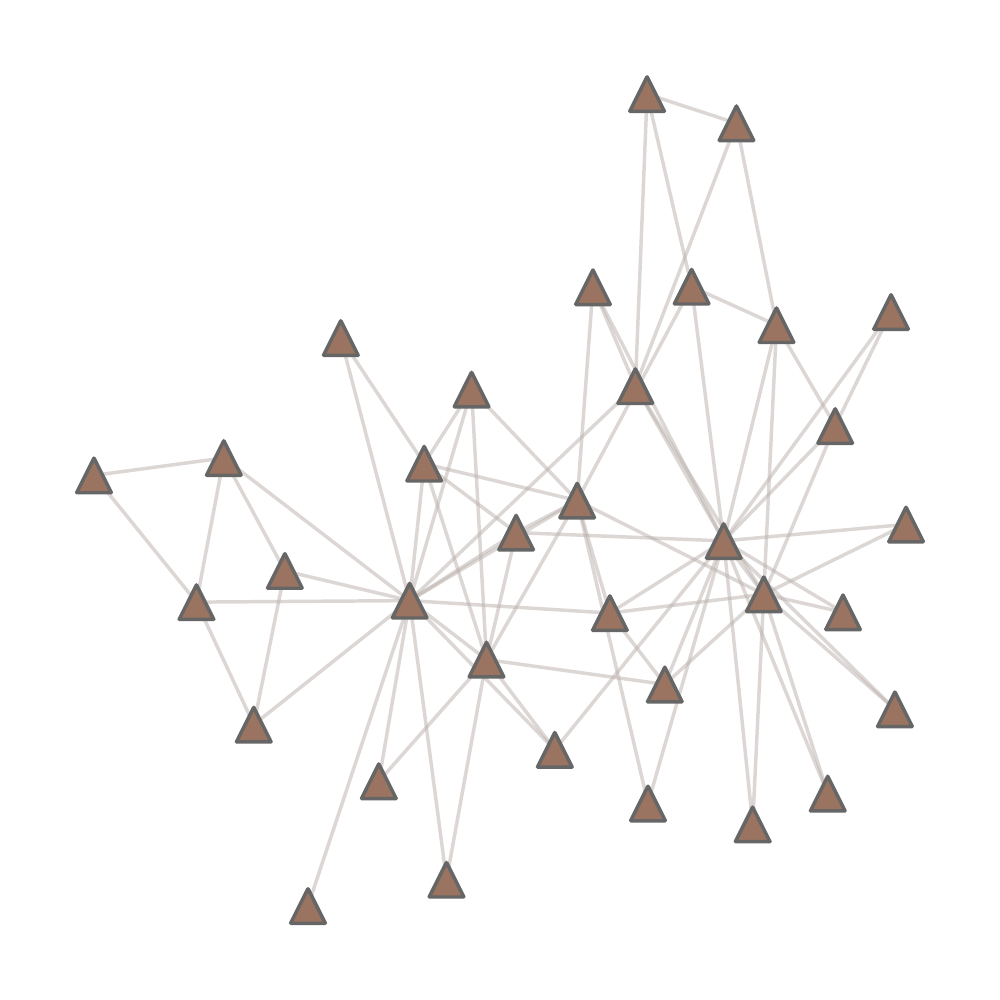}
        \label{fig:karate}
    }
    \hspace*{-6pt}
    \subfloat[The result of i-cNRL]{
        \includegraphics[width=0.5\linewidth]{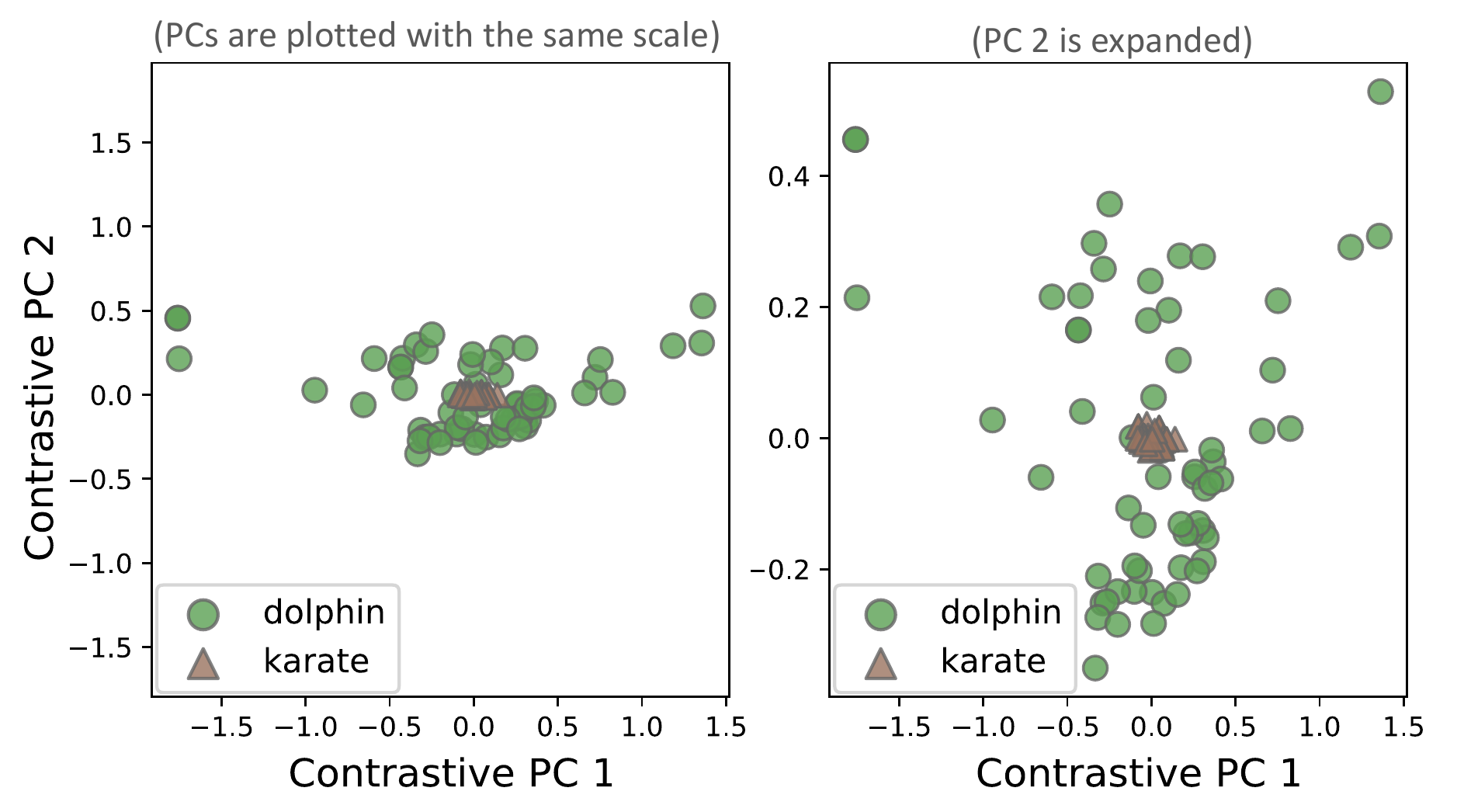}
        \label{fig:dolphin_karate_cpca}
    }
    \\
    \hspace*{-10pt}
    \subfloat[$\GraphT$: Dolphin]{
        \includegraphics[width=0.23\linewidth]{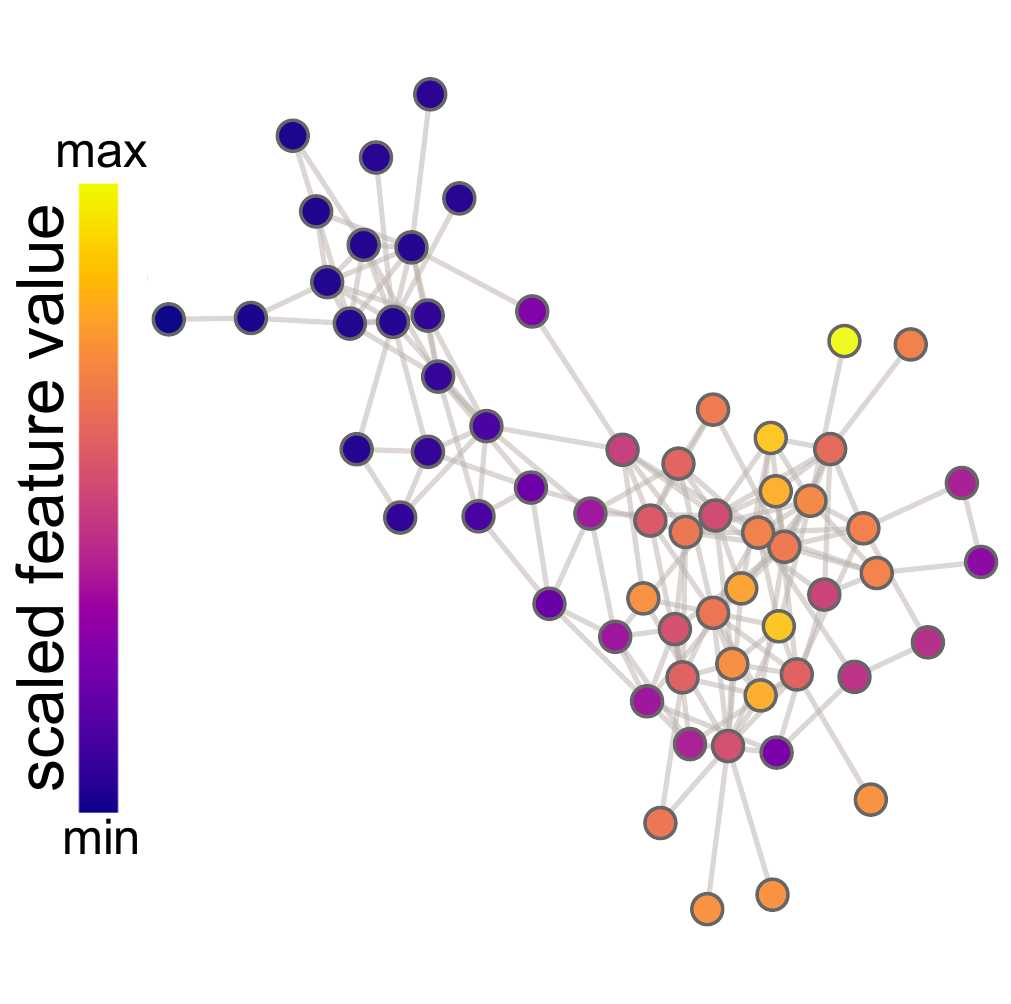}
        \label{fig:dolphin_colored}
    }
    \hspace*{-10pt}
    \subfloat[$\GraphB$: Karate]{
        \includegraphics[width=0.23\linewidth]{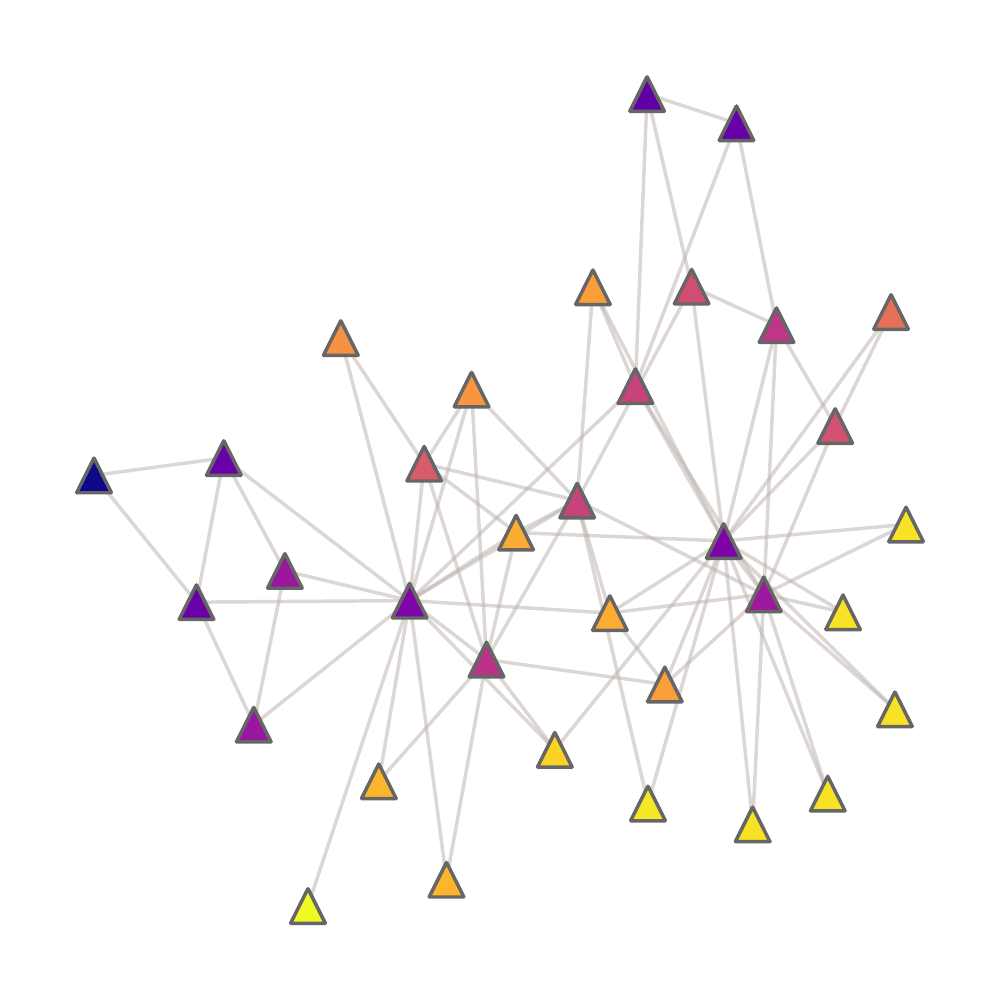}
        \label{fig:karate_colored}
    }
    \hspace*{-6pt}
    \subfloat[The result of i-cNRL]{
        \includegraphics[width=0.5\linewidth]{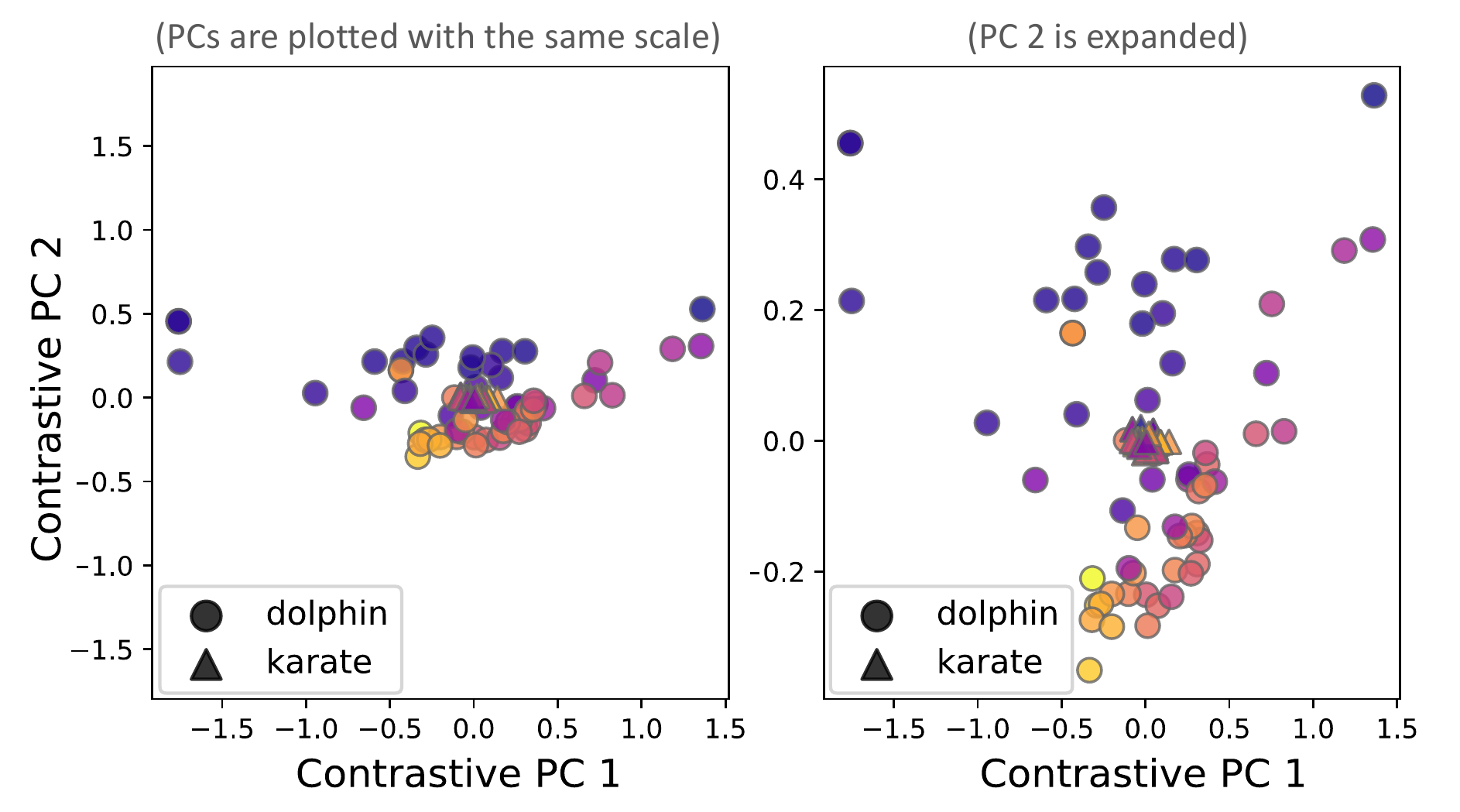}
        \label{fig:dolphin_karate_cpca_colored}
    }
    \caption{
    (a) and (b) show the dolphin social network and the Zachary's karate club network, used as $\GraphT$ and $\GraphB$ for i-cNRL, respectively.
    (c) shows the i-cNRL results with 2D embedding.
    (d), (e), and (f) colorcode each node in (a), (b), and (c) based on the top-contributed feature (\texttt{\DolphinVsKarate-10}) of the first contrastive principal component (\texttt{cPC~1}): $(\Phi_{\rm mean})(\rm \bf x)$ with `eigenvector' as the base feature $\rm \bf x$ (see \autoref{table:dolphin_karate_loadings}). 
    }
	\label{fig:dolphin_vs_karate}
\end{figure}

To provide an illustrative example of analysis with cNRL, we begin by comparing two social networks. 
We use the Dolphin social network~\citep{lusseau2003bottlenose} as $\GraphT$ and the Zachary's karate club network~\citep{zachary1977information} as $\GraphB$. 
\autoref{fig:dolphin_vs_karate}\subref{fig:dolphin} and \subref{fig:karate} depict the network structures of these networks. 
The statistics of these networks can be found in \autoref{table:network_info} (see \texttt{\DolphinNwID} and \texttt{\KarateNwID}).
By comparing these two networks, we want to reveal unique patterns in the Dolphin social network and identify which network characteristics relate to the patterns. 

\begin{table}[tb]
    \centering
    \footnotesize
    \caption{Statistics of network datasets.}
    \label{table:network_info}
    \input{tables/network_info.tex}
\end{table}

We apply our i-cNRL to the two networks and then plot a 2D embedding result with contrastive PCA (cPCA)~\citep{abid2018exploring}, as shown in 
\autoref{fig:dolphin_vs_karate}\subref{fig:dolphin_karate_cpca}. 
The $x$- and $y$-directions in \autoref{fig:dolphin_vs_karate}\subref{fig:dolphin_karate_cpca} represent the first and second contrastive principal components (cPCs), respectively. 
Details of i-cNRL and related techniques will be described in \autoref{sec:methodology}.
\autoref{fig:dolphin_vs_karate}\subref{fig:dolphin_karate_cpca} shows that the nodes in $\GraphT$ are more widely distributed, whereas the nodes in $\GraphB$ are placed only around the center, which reveals some patterns specific to $\GraphT$ when compared with $\GraphB$.

\begin{table}[tb]
    \centering
    \footnotesize
    \caption{Learned features and their cPC loadings for the Dolphin vs. Karate example.}
    \label{table:dolphin_karate_loadings}
    \input{tables/dolphin_karate_loadings}
\end{table}

Moreover, since i-cNRL offers interpretability to the learned results, we can analyze why the above patterns appear.
As shown in \autoref{table:dolphin_karate_loadings}, the method provides loadings of each cPC (or cPC loadings), of which the absolute value indicates how large each learned feature contributes to each cPC direction.
Each learned feature can be represented as a combination of the relational function $\RelFunc$ and the base feature $\BaseFeat$ (see \autoref{sec:methodology} for details). 
\autoref{table:dolphin_karate_loadings} indicates that feature \texttt{\DolphinVsKarate-10} has the highest contribution to \texttt{cPC 1}.
From the relational function $(\RelFeatOpe{}{\Mean})(\BaseFeat)$ and the base feature `eigenvector'~\citep{newman2018networks}, this feature is interpreted as ``the mean eigenvector centrality of the neighbors of a node.''

To investigate the relationships between this feature and the i-cNRL result, we colorcode the network nodes in \autoref{fig:dolphin_vs_karate}\subref{fig:dolphin}, \subref{fig:karate}, and \subref{fig:dolphin_karate_cpca} based on the feature values, as shown in \autoref{fig:dolphin_vs_karate}\subref{fig:dolphin_colored}, \subref{fig:karate_colored}, and \subref{fig:dolphin_karate_cpca_colored}. 
Here, the feature values are scaled with the standardization when applying i-cNRL.
We can see that, in \autoref{fig:dolphin_vs_karate}\subref{fig:dolphin_karate_cpca_colored} (right), the nodes around the top-left corner tend to have smaller feature values while the nodes around the bottom-right tend to have higher values. 
By comparing with \autoref{fig:dolphin_vs_karate}\subref{fig:dolphin_colored}, we notice that these two node groups correspond to the top-left and bottom-right communities in \autoref{fig:dolphin_vs_karate}\subref{fig:dolphin_colored}.  
Since the feature value shows the mean eigenvector centrality of the neighbors of a node, the nodes in the top-left community tend to have a low eigenvector centrality including their neighbors. 
On the other hand, the nodes in the right-bottom community have neighbors with a high eigenvector centrality.
\autoref{fig:dolphin_vs_karate}\subref{fig:karate_colored} indicates that $\GraphB$ does not have such clearly separated communities by the feature values, unlike $\GraphT$. 
Therefore, i-cNRL learns the patterns highly related to the eigenvector centralities of each node's neighbors, which can clearly separate the two communities in the Dolphin social network. 

%% file: tables/network_info.tex
\begin{tabular}{rlrrl}
\toprule
 ID &                                                                 Name &  \# of nodes &  \# of links &  Directed \\
\midrule
  \texttt{\DolphinNwID} &                                 Dolphin~\citep{lusseau2003bottlenose} &           62 &          159 &     False \\
  \texttt{\KarateNwID} &                                 Karate~\citep{zachary1977information} &           34 &           78 &     False \\
  \texttt{\RandomNwID} &                                                               Random &          100 &          471 &      True \\
  \texttt{\PriceNwID} &                                                                Price &          100 &          294 &      True \\
  \texttt{\PtoPNwID} &  p2p-Gnutella08~\citep{ripeanu2002mapping} &         6,301 &        20,777 &      True \\
  \texttt{\PriceTwoNwID} &                                                              Price 2 &         6,301 &        18,897 &      True \\
  \texttt{\EPriceNwID} &                                                       Enhanced Price &         6,301 &        18,281 &      True \\
  \texttt{\CombinedAPMSNwID} &                   Combined-AP/MS~\citep{collins2007toward} &         1,622 &         9,070 &     False \\
  \texttt{\LCMultiNwID} &                LC-multiple~\citep{reguly2006comprehensive} &         1,536 &         2,925 &     False \\
  \texttt{\SchFirstDayNwID} &                         School-Day1~\citep{stehle2011high} &          236 &         5,899 &     False \\
  \texttt{\SchSecondDayNwID} &                         School-Day2~\citep{stehle2011high} &          238 &         5,539 &     False \\
\bottomrule
\end{tabular}

%% file: tables/dolphin_karate_loadings.tex
\begin{tabular}{rllrr}
\toprule
 ID &                                          relational function $\RelFunc$ & base feature $\BaseFeat$ &  \texttt{cPC 1} &  \texttt{cPC 2} \\
\midrule
  \texttt{\DolphinVsKarate-1} &                                                           $(\BaseFeat)$ &             total-degree &   0.01 &  -0.09 \\
  \texttt{\DolphinVsKarate-2} &                                                           $(\BaseFeat)$ &              betweenness &  -0.02 &  -0.02 \\
  \texttt{\DolphinVsKarate-3} &                                                           $(\BaseFeat)$ &                closeness &   0.01 &   0.01 \\
  \texttt{\DolphinVsKarate-4} &                                                           $(\BaseFeat)$ &              eigenvector &  -0.11 &   0.00 \\
  \texttt{\DolphinVsKarate-5} &                                                           $(\BaseFeat)$ &                 PageRank &   0.11 &   0.18 \\
  \texttt{\DolphinVsKarate-6} &                                                           $(\BaseFeat)$ &                     Katz &   0.01 &  -0.08 \\
  \texttt{\DolphinVsKarate-7} &                                 $(\RelFeatOpe{}{\rm \Mean})(\BaseFeat)$ &             total-degree &  -0.18 &  -0.42 \\
  \texttt{\DolphinVsKarate-8} &                                 $(\RelFeatOpe{}{\rm \Mean})(\BaseFeat)$ &              betweenness &   0.15 &  -0.04 \\
  \texttt{\DolphinVsKarate-9} &                                 $(\RelFeatOpe{}{\rm \Mean})(\BaseFeat)$ &                closeness &  -0.24 &   0.03 \\
  \texttt{\DolphinVsKarate-10} &                                 $(\RelFeatOpe{}{\rm \Mean})(\BaseFeat)$ &              eigenvector &   \textbf{0.80} &   0.10 \\
  \texttt{\DolphinVsKarate-11} &                                 $(\RelFeatOpe{}{\rm \Mean})(\BaseFeat)$ &                 PageRank &  -0.35 &   \textbf{0.76} \\
  \texttt{\DolphinVsKarate-12} &                                 $(\RelFeatOpe{}{\rm \Mean})(\BaseFeat)$ &                     Katz &  -0.25 &  -0.43 \\
  \texttt{\DolphinVsKarate-13} &                                  $(\RelFeatOpe{}{\rm \Max})(\BaseFeat)$ &                 PageRank &  -0.02 &   0.00 \\
  \texttt{\DolphinVsKarate-14} &  $(\RelFeatOpe{}{\rm \Mean} \circ \RelFeatOpe{}{\rm \Mean})(\BaseFeat)$ &             total-degree &  -0.17 &  -0.01 \\
  \texttt{\DolphinVsKarate-15} &   $(\RelFeatOpe{}{\rm \Mean} \circ \RelFeatOpe{}{\rm \Max})(\BaseFeat)$ &                 PageRank &   0.02 &  -0.00 \\
\bottomrule
\end{tabular}

%% file: 4_short_methodology.tex
\begin{figure}[tb]
	\centering
    \includegraphics[width=1.0\linewidth]{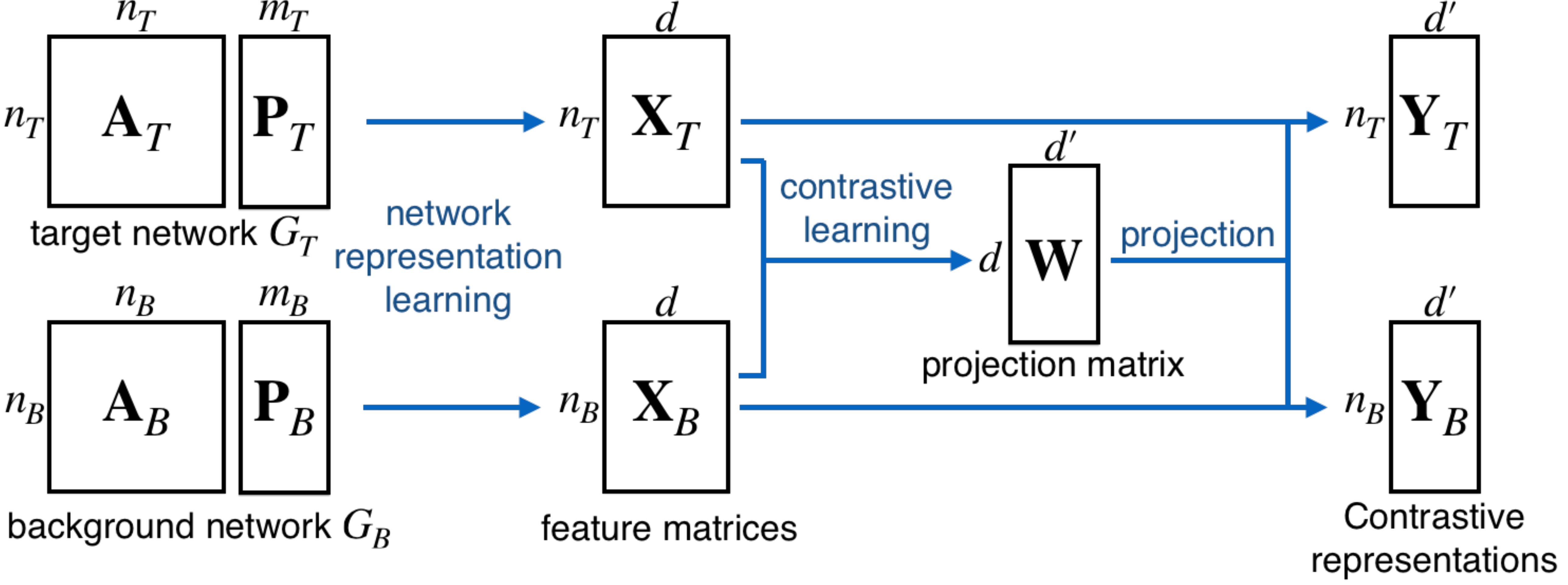}
    \caption{The general architecture for cNRL.}
	\label{fig:architecture}
\end{figure}

\section{cNRL Architecture}
\label{sec:architecture}
\autoref{fig:architecture} shows a general architecture for cNRL. 
Notations used for the following sections are listed in \autoref{table:notation}.
The current contrastive learning methods~\citep{zou2013contrastive,abid2018exploring,dirie2019contrastive,abid2019contrastive,severson2019unsupervised} require target and background feature matrices ($\FeatMatT$ and $\FeatMatB$) sharing the same features as inputs.
However, matrices that represent target and background networks ($\GraphT$ and $\GraphB$) such as adjacency matrices ($\AdjT$ and $\AdjB$) might have a different number of nodes or no correspondence in nodes of $\AdjT$ and $\AdjB$.
Thus, we cannot directly apply the contrastive learning methods to $\GraphT$ and $\GraphB$. 
To address this issue, our core idea of cNRL consists of two main steps: (1) generating feature matrices $\FeatMatT$ and $\FeatMatB$ from networks $\GraphT$ and $\GraphB$ respectively by using NRL, and (2) applying contrastive learning on $\FeatMatT$ and $\FeatMatB$. 
Also, we want to emphasize that cNRL cannot be achieved by simply combining NRL and contrastive learning. 
For example, a method for NRL needs to satisfy a certain requirement to enable contrastive learning in the ensuing step. 
We identify such requirements for cNRL.

Below we describe the details of each part of the cNRL architecture with requirements on inputs, NRL, and contrastive learning methods.
Here we focus only on node feature learning to provide a simple and clear explanation. 
However, the architecture is generic enough to be used for link (or edge) feature learning.

\begin{table}[tb]
    \centering
    \footnotesize
    \caption{Summary of notation.}
    \label{table:notation}
    \input{tables/notation.tex}
\end{table}

\vspace{2pt}
\noindent\textbf{Inputs.}
cNRL takes $\GraphT$ and $\GraphB$ as inputs. 
These networks can be any combination of being undirected or directed, unweighted or weighted, and non-attributed or attributed. 
The numbers of $\GraphT$ and $\GraphB$ nodes (i.e., $\nNodesT$ and $\nNodesB$) do not have to be the same. 
Similarly, the numbers of attributes $\nAttrsT$ and $\nAttrsB$ may be different. 

\vspace{2pt}
\noindent\textbf{Network representation learning.}
The first step in \autoref{fig:architecture} is applying an NRL method in order to transform the inputs $\GraphT$ and $\GraphB$ to feature matrices $\FeatMatT$ and $\FeatMatB$, respectively. 
Contrastive learning requires that $\FeatMatT$ and $\FeatMatB$ share the same features by nature of its learning purpose. 
Therefore, for this process, we need to use an NRL method that can produce the same features across networks.

\vspace{2pt}
\noindent\textbf{Contrastive learning.}
Once we obtain $\FeatMatT$ and $\FeatMatB$, which have the same $\nNRLFeats$ learned features, we can apply any of the contrastive learning methods using $\FeatMatT$ and $\FeatMatB$ as target and background datasets, respectively. 
Contrastive learning generates a parametric mapping (or a projection matrix $\ProjMat$) from $\nNRLFeats$ features learned by NRL to $\nCNRLFeats$ contrastive features ($\nCNRLFeats \leq \nNRLFeats$).
With this projection matrix, $\FeatMatT$ and $\FeatMatB$ can be transformed to contrastive representations $\ContReprT$ and $\ContReprB$, respectively. 
As the existing contrastive learning works~\citep{zou2013contrastive,abid2018exploring,dirie2019contrastive,abid2019contrastive,severson2019unsupervised} only produced $\ContReprT$ for their analysis, the generation of $\ContReprB$ is optional. 
However, as demonstrated in \autoref{fig:dolphin_vs_karate}\subref{fig:dolphin_karate_cpca}, by visualizing both $\ContReprT$ and $\ContReprB$ in one plot, we can clearly see whether contrastive learning has found unique patterns in $\GraphT$ relative to $\GraphB$.
This visualization is a new approach to understanding contrastive learning results, where we can judge whether or not contrastive learning successfully finds target dataset's (in cNRL, target network's) unique patterns.

\section{Interpretable cNRL Method}
\label{sec:methodology}

As a specific method using the architecture above, we describe i-cNRL, which employs DeepGL~\citep{rossi2018deep} for NRL and cPCA~\citep{abid2018exploring} for contrastive learning.
This algorithm selection is vital to provide interpretability; thus, we also provide the design rationale for the selection.

\subsection{Network Representation Learning} 

As stated in \autoref{sec:architecture}, NRL needs to generate $\FeatMatT$ and $\FeatMatB$ that have the same features.
To achieve this, we can employ any \emph{inductive} NRL method~\citep{rossi2018deep} (e.g., GraphSAGE by \citealp{hamilton2017inductive} and FastGCN by \citealp{chen2018fastgcn}). 
However, we want to provide the interpretability in the contrastive representations obtained by cNRL; thus, an NRL method needs to generate interpretable features as the learned result.
As a result, we specifically use DeepGL~\citep{rossi2018deep} in the first step of i-cNRL. 

\subsubsection{DeepGL}

DeepGL learns node and link features consisting of the \textit{base feature} $\BaseFeat$ and \textit{relational function} $\RelFunc$. 
For a concise explanation, we describe DeepGL for only node feature learning. 

A base feature $\BaseFeat$ is any simple feature or measure we can obtain for each node. 
For example, $\BaseFeat$ can be (weighted) in-, out-, total-degree, degeneracy (or $k$-core numbers), PageRank~\citep{newman2018networks}, or a node attribute (e.g., gender of a node in a social network).

A relational function $\RelFunc$ is a combination of \textit{relational feature operators}, which is applied to a base feature.
A relational feature operator summarizes base feature values of one-hop neighbors of a node. 
For example, the operator can be a computation of the mean, sum, maximum base feature values of one-hop neighbors' of a node.
Also, the neighbors can be either in-, out-, total-neighbors. 
Together with the summary measure $S$ (e.g., mean), the operators can be denoted $\RelFeatOpe{-}{S}$, $\RelFeatOpe{+}{S}$, and $\RelFeatOpe{}{S}$, respectively. 
For example, $\RelFeatOpe{-}{\Mean}(\BaseFeat)$ computes the mean $\BaseFeat$ of the in-neighbors of a node. 
Moreover, the relational feature operator can be applied repeatedly. 
For example, $\RelFunc = (\RelFeatOpe{+}{\Mean} \circ \RelFeatOpe{-}{\Max})(\BaseFeat)$ first computes the maximum $\BaseFeat$ of in-neighbors for each out-neighbor of a node and then produces the mean of these maximum values. 
As described with the examples above, $\BaseFeat$ and $\RelFunc$ are combinations of simple measures and operators; thus, both are interpretable.

In DeepGL, we can select as many different base features and relational feature operators as we want to consider. 
The learning process contains $h$ number of iterations (indicated by the user), and in the end we obtain all the learned features $\F = \{\F_0, \F_1, \cdots, \F_\DepthRelFunc\}$, each of which is a relational function over a base feature $\RelFunc(\BaseFeat)$.
During each iteration, DeepGL prunes redundant features based on the similarities of the obtained feature values (refer to \citealt{rossi2018deep} for details).
\autoref{table:dolphin_karate_loadings} shows an example of learned features from the Dolphin social network~\citep{lusseau2003bottlenose}.

\subsubsection{Use of Transfer Learning with DeepGL for cNRL}

As described above, the learned features $\F$ by DeepGL are the combinations of the base features and relational functions. 
Once we obtain $\F$ from one network, we can naturally compute $\F$ for other networks.
That is, DeepGL is inductive and can be used for transfer learning~\citep{rossi2018deep}. 

In cNRL, we need to decide which network(s), $\GraphT$ and/or $\GraphB$, should be used for learning $\F$. 
One possible choice is applying DeepGL for both to learn the features of target and background networks ($\F_T$ and $\F_B$, respectively). Then, we can use the union of these features (i.e., $\F_T \cup \F_B$) for producing feature matrices $\FeatMatT$ and $\FeatMatB$. 
However, since cNRL aims to identify unique patterns in $\GraphT$ relative to $\GraphB$, such as patterns where only $\GraphT$ has high variance (see \autoref{sec:icnrl_cl}), only a set of features capturing $\GraphT$'s characteristics is required. 
Thus, we apply DeepGL to $\GraphT$ and use the learned features $\F_T$ for both $\GraphT$ and $\GraphB$ to generate $\FeatMatT$ and $\FeatMatB$.
It can also avoid unnecessary computation for learning $\F_B$ from $\GraphB$. 

\subsection{Contrastive Learning}
\label{sec:icnrl_cl}
The above NRL step generates feature matrices $\FeatMatT$ and $\FeatMatB$. 
The remaining step is learning contrastive representations $\ContReprT$ and $\ContReprB$ through contrastive learning. 
While we can use any contrastive learning method, one of our goals is to provide interpretability.
Since DeepGL generates interpretable features for $\FeatMatT$ and $\FeatMatB$, we can provide interpretable $\ContReprT$ and $\ContReprB$ by using a method that reveals interpretable relationships between $\nNRLFeats$ features learned by NLR and $\nCNRLFeats$ features learned by contrastive learning. 
Among current contrastive learning methods~\citep{zou2013contrastive,abid2018exploring,dirie2019contrastive,abid2019contrastive,severson2019unsupervised}, only cPCA or its variants (e.g., sparse cPCA~\citep{boileau2020exploring}) can provide such relationships by utilizing the linearity of its algorithm in a similar manner to ordinary PCA~\citep{jolliffe1986principal}. 
Thus, we select cPCA for the second step of i-cNRL, though, it can be replaced with any other interpretable contrastive learning methods developed in the future.

\subsubsection{Contrastive PCA (cPCA)}

cPCA~\citep{abid2018exploring} is a variant of PCA for contrastive learning. 
Similar to the classical PCA, cPCA first applies centering to each feature of $\FeatMatT$ and $\FeatMatB$ and then obtains their corresponding covariance matrices $\CovT$ and $\CovB$.
Let \(\mathbf{v}\) be any unit vector of $d$ length. 
Then, with a given direction $\mathbf{v}$, the variances for $\FeatMatT$ and $\FeatMatB$ can be written as:
$\smash{\sigma_{T}^2(\mathbf{v}) \overset{\scriptscriptstyle\mathrm{def}}{=} \mathbf{v}^{\mathsf{T}} \CovT \mathbf{v}}$, $\smash{ 
\sigma_{B}^2(\mathbf{v}) \overset{\scriptscriptstyle\mathrm{def}}{=} \mathbf{v}^{\mathsf{T}} \CovB \mathbf{v}}$.
The optimization that finds a direction $\smash{\mathbf{v}^*}$ where $\FeatMatT$ has high variance but $\FeatMatB$ has low variance can thus be written as:
\begin{equation}
\label{eq:contrastive_direction}
     \mathbf{v}^* = \operatorname*{argmax}_{\mathbf{v}} ~ \sigma_{T}^2(\mathbf{v}) - \alpha \sigma_{B}^2(\mathbf{v}) = \operatorname*{argmax}_{\mathbf{v}} ~ \mathbf{v}^{\mathsf{T}} (\CovT - \ContParam \CovB) \mathbf{v}
\end{equation}
where $\ContParam$ is a contrast parameter ($0 \leq \ContParam \leq \infty$). 
From \autoref{eq:contrastive_direction}, we can see that $\smash{\mathbf{v}^*}$ corresponds to the first eigenvector of the matrix $\smash{\mathbf{C} \overset{\scriptscriptstyle\mathrm{def}}{=} (\CovT - \ContParam \CovB)}$.
The eigenvectors of $\mathbf{C}$ can be calculated with eigenvalue decomposition. 
These computed eigenvectors are called contrastive principal components (cPCs) and are orthogonal to each other. 
Similar to the classical PCA, we can obtain top-$\nCNRLFeats$ cPCs as the learned features. 
With projection matrix $\ProjMat$ consisting of $\nCNRLFeats$ cPCs (i.e., $\ProjMat$ is a $\nNRLFeats \times \nCNRLFeats$ matrix), we can obtain the contrastive representation $\ContReprT$ of $\FeatMatT$.

\begin{figure}[tb]
	\centering
	\captionsetup{farskip=0pt} 
    \includegraphics[width=1.0\linewidth]{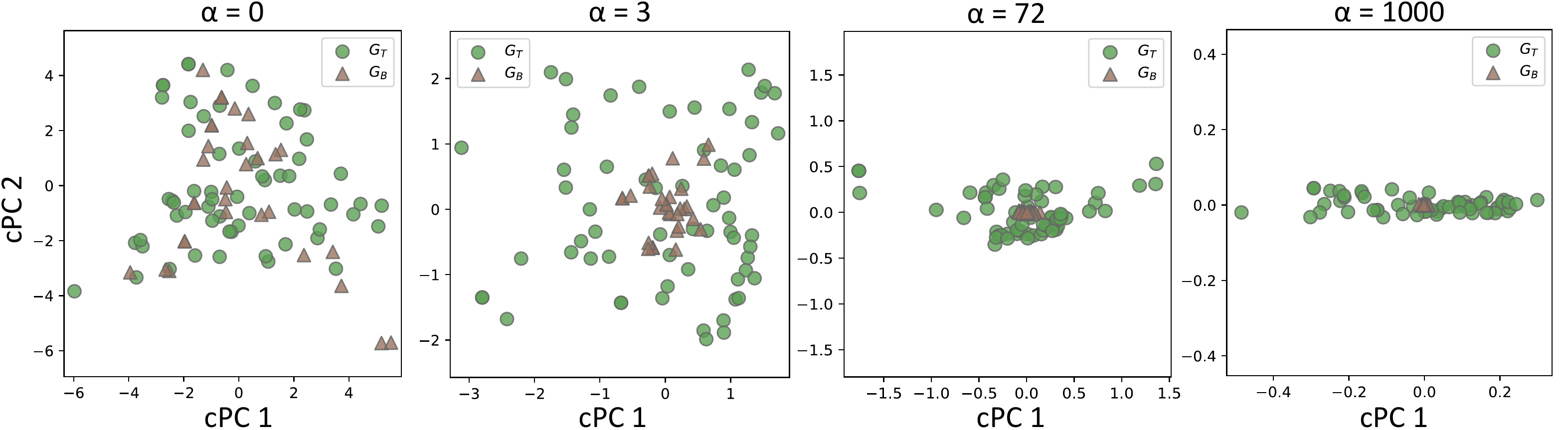}
    \caption{
    The cPCA results with different $\ContParam$ values, applied on feature matrices $\FeatMatT$ and $\FeatMatB$ generated from $\GraphT$ (the dolphin network) and $\GraphB$ (the Karate network) in \autoref{fig:dolphin_vs_karate}.
    When $\ContParam = 0$, the result is the same with applying PCA on $\FeatMatT$. A decrease of $\FeatMatB$'s variances is observed as $\ContParam$ increases. The result with $\ContParam = 72$ corresponds to the results in \autoref{fig:dolphin_vs_karate}.
    }
	\label{fig:multiple_alphas}
\end{figure}

The above contrast parameter $\alpha$ controls the trade-off between having high target variance and low background variance. 
When $\ContParam = 0$, cPCs only maximize the variance of $\FeatMatT$, the same as those in the classical PCA.
As $\ContParam$ increases, cPCs place greater emphasis on directions that reduce the variance of $\FeatMatB$.
\autoref{fig:multiple_alphas} shows the results of cPCA with different $\ContParam$ values. 
Because $\ContParam$ has a strong impact on the result, \citet{abid2018exploring} introduced the semi-automatic selection of $\ContParam$ utilizing spectral clustering~\citep{ng2002spectral}. 
We go one step further to provide a fully automatic selection of $\ContParam$ (see Automatic Contrast Parameter Selection explained below).

\subsubsection{Representation Learning with cPCA in cNRL}

By applying cPCA to $\FeatMatT$ and $\FeatMatB$, we can generate the projection matrix $\ProjMat$ and contrastive representations $\ContReprT$ and $\ContReprB$. 
Because each learned feature by DeepGL could have a different scale, as a default, our method applies the standardization (i.e., scaling each feature to have zero mean and unit variance).
Also, there could be a scale difference between $\FeatMatT$ and $\FeatMatB$ (e.g., $\GraphT$ and $\GraphB$'s mean total-degrees are 1,000 and 100, respectively). 
Thus, instead of a concatenated matrix of $\FeatMatT$ and $\FeatMatB$, we apply the standardization to each of $\FeatMatT$ and $\FeatMatB$ for both learning and projection. 
This approach can avoid generating a 2D embedding result where $\GraphT$'s nodes are placed extremely far away from $\GraphB$'s nodes.
In addition, the selection of $\alpha$ becomes easier.
For example, without the above standardization, $\FeatMatT$ could have a much smaller variance in any direction $\mathbf{v}$ when compared to $\FeatMatB$; consequently, to find unique patterns, we may need to set excessively small $\alpha$ (e.g., $\alpha=0.0001$), and vice versa.
On the other hand, when the comparison of the absolute value differences between $\FeatMatT$ and $\FeatMatB$ is important, we can apply the standardization to the concatenated matrix.

To provide interpretable relationships between NLR features $\nNRLFeats$ and contrastive learning features $\nCNRLFeats$, 
we extract contrastive PC loadings (cPC loadings), which are coefficients of $\ProjMat$\footnote{Similar to ordinary PCA, instead of loadings, we could also refer to loadings scaled with a square root of the corresponding eigenvalue. In ordinary PCA, the scaled loadings show the correlations between each feature and the corresponding PC.
However, cPCA's scaled loadings do not show such correlations as cPCA performs eigenvalue decomposition on ($\CovT - \ContParam \CovB$), instead of $\CovT$ or $\CovB$.}
These cPC loadings indicate how strongly each of the $d$ input features contributes to the corresponding cPC. 
\autoref{table:dolphin_karate_loadings} shows an example of cPC loadings for the first and second cPCs.
As demonstrated in \autoref{sec:analysis_example}, by referring to a list of the learned features via NRL and cPC loadings, we can interpret the obtained representations $\ContReprT$ and $\ContReprB$.

Also, as discussed in \autoref{sec:architecture}, we visualize both $\ContReprT$ and $\ContReprB$ in one plot to judge whether or not i-cNRL has found unique patterns in $\GraphT$. 
cPCA contrasts variances of $\ContReprT$ and $\ContReprB$; thus, when $\GraphT$ has unique patterns, $\ContReprT$ has a much higher variance than $\ContReprB$  (e.g., when $\ContParam=72$ and $\ContParam=1000$ in \autoref{fig:multiple_alphas}).
Note that, in such a case, $\GraphB$'s nodes tend to be highly overlapped with each other in a visualized result, and it becomes difficult to see the difference of each of $\GraphB$'s nodes.
However, this is preferable for analyses using i-cNRL as the visualization of the embedding result should highlight unique patterns in $\GraphT$ but not the differences within $\GraphB$'s nodes.

\subsubsection{Automatic Contrast Parameter Selection}
We now show how to automatically select the parameter $\alpha$ in cPCA. Since we want to maximize the variation in the target feature matrix while simultaneously minimizing the variation in the background feature matrix, we can solve the following ratio problem:
\begin{align}
\label{eq:ratio}
    \max_{\ProjMat^\top \ProjMat = I_\nCNRLFeats} ~ \frac{\tr(\ProjMat^\top \CovT \ProjMat)}{\tr(\ProjMat^\top \CovB \ProjMat)}.
\end{align}
While directly solving \autoref{eq:ratio} may be difficult, there is a convenient iterative algorithm due to \citet{Dinkelbach67}. The algorithm consists of two steps. Given $\ProjMat_t$, we perform
\begin{itemize}
    \item $\alpha_t \gets \dfrac{\tr(\ProjMat^\top_t \CovT \ProjMat_t)}{\tr(\ProjMat^\top_t \CovB \ProjMat_t)}$\\
    \item $\ProjMat_{t+1} \gets \argmax\limits_{\ProjMat^\top \ProjMat = I_\nCNRLFeats} ~ \tr(\ProjMat^\top (\CovT-\alpha_t\CovB) \ProjMat)$.
\end{itemize}
Clearly, $\alpha_t$ is just the objective value of our ratio problem, \autoref{eq:ratio}, evaluated at the current solution $\ProjMat_t$. 
$\alpha_t$ monotonically increases to the maximum value, and the convergence is usually very quick (e.g., less than 10 iterations, as evaluated in \autoref{app:auto_alpha}). Conveniently, the second step for finding the next solution $\ProjMat_{t+1}$ is just the original cPCA problem, where we use $\alpha_t$ as our trade-off parameter. We can also regard cPCA as a one-shot algorithm for the ratio problem, \autoref{eq:ratio}, where the user specifies $\alpha$. 
One problem of the method above is that $\ContParam_t$ reaches close to infinite when $\CovB$ is nearly singular.
In this case, the algorithm finds an embedding where $\CovT$
has nonzero variance while $\CovB$ has zero variance.
To avoid this problem, our method simply adds a small constant value $\ContConst$, as a default $\ContConst = 10^{-3}$, to each diagonal element of $\CovB$. 
As discussed, by default, $\FeatMatB$ is standardized and each diagonal element $\CovB$ is close to one; thus, $\ContConst = 10^{-3}$ is reasonably small to avoid introducing a strong bias.
We note that the above algorithm of \citet{Dinkelbach67} has been used in discriminant analysis \citep{GuoLYSW03,JiaNZ09}, whose motivation is different from ours.

\subsection{Complexity Analysis}

The time and space complexities of i-cNRL are comparable to those of DeepGL and cPCA.
According to \citet{rossi2018deep}, DeepGL's time and space complexities for learning from $\GraphT$ are $\mathcal{O}(d(\nEdgesT + d\nNodesT))$ and $\mathcal{O}(d\nNodesT)$, respectively, where $\nEdgesT$ is the number of links in $G_T$. 
Note that the time and space complexities for computing base features are assumed lower than these. 
When including the transfer learning step to obtain $\FeatMatB$, the space complexity becomes $\mathcal{O}(d(\nNodesT + \nNodesB))$. 
For a fixed $\ContParam$, cPCA has the similar time and space complexities with PCA, which are $\mathcal{O}(d^2 (\nNodesT + \nNodesB) + d^3))$ and $\mathcal{O}(d^2)$. 
Even with the automatic selection of $\ContParam$ in \autoref{sec:icnrl_cl}, we can assume that these complexities stay the same.
This is because the automatic selection usually only needs a small number of iterations and does not require storing of additional information.
Thus, in total, i-cNRL has the time complexity $\mathcal{O}(d(\nEdgesT + d(\nNodesT + \nNodesB) + d^2))$ and the space complexity $\mathcal{O}(d(\nNodesT + \nNodesB + d))$. 
However, in practice, $d$, the number of features learned by NRL, should be much smaller than the numbers of nodes and links of $G_T$ and $G_B$.
Under this assumption, the time and space complexities are $\mathcal{O}(d(\nEdgesT + d(\nNodesT + \nNodesB)))$ and $\mathcal{O}(d(\nNodesT + \nNodesB))$, respectively. 
This indicates that the computational cost is largely due to DeepGL.

The above fact has guided our finer level design choices of i-cNRL.
To reduce the computational cost of DeepGL, we utilize DeepGL's feature pruning, instead of keeping all features for cPCA.
However, if fast computation is not required, we could skip feature pruning and apply sparse cPCA~\citep{boileau2020exploring}.
Sparse cPCA uses a limited number of features when constructing cPCs (i.e., $\ProjMat$ becomes a sparse matrix), and thus helps produce interpretable results from many features.
Also, we could choose another design alternative to more tightly connect cPCA with feature pruning in DeepGL. 
For example, instead of the similarities of features, we can prune features by checking whether or not an inclusion of the corresponding feature highly influences the optimization result with \autoref{eq:contrastive_direction}.
However, this design also requires more computations as the pruning needs to derive feature values for both $\GraphT$ and $\GraphB$ and apply cPCA repeatedly during the learning process of DeepGL.

To provide better computational scalability, DeepGL can utilize parallelization. 
When learning $\F_i$ (i.e., features with $i$ relational feature operators), we can individually apply relational feature operators to each graph element (e.g., graph node) and/or each base feature.
For details, refer to the evaluation performed by \cite{rossi2018deep}.

%% file: tables/notation.tex
\begin{tabular}{rl}
\toprule
\multicolumn{2}{l}{
\bgcolored{\textbf{Notations for cNRL}}{0.73\linewidth}} \\
$\GraphT$, $\GraphB$ & target and background networks \\
$\AdjT$, $\AdjB$ & adjacency matrices of $\GraphT$ and $\GraphB$ \\
$\AttrT$, $\AttrB$ & matrices of node attributes of $\GraphT$ and $\GraphB$ \\
$\nNodesT$, $\nNodesB$ & numbers of nodes in $\GraphT$ and $\GraphB$ \\
$\nAttrsT$, $\nAttrsB$ & numbers of attributes in $\GraphT$ and $\GraphB$ \\
$\nEdgesT$, $\nEdgesB$ & numbers of edges in $\GraphT$ and $\GraphB$ \\
$\nNRLFeats$,  $\nCNRLFeats$ & numbers of features learned by NRL and contrastive learning\\
$\FeatMatT$, $\FeatMatB$ & target and background feature matrices \\
$\ProjMat$ & projection matrix learned by contrastive learning \\
$\ContReprT$, $\ContReprB$ & contrastive representations of $\FeatMatT$ and $\FeatMatB$ \\
\noalign{\vskip 5pt} 
\multicolumn{2}{l}{
\bgcolored{\textbf{Notations for DeepGL}}{0.73\linewidth}} \\
$\BaseFeat$ & base feature (e.g., in-degree) \\
$\RelFunc$ & relational function \\
$\RelFeatOpe{-}{}$, $\RelFeatOpe{+}{}$, $\RelFeatOpe{}{}$ & relational feature operators for in-, out-, total neighbors\\
$\SummaryMeasure$ & summary measure (e.g., mean, sum, and maximum) \\
$\F_i$ & set of learned features with $i$ relational feature operators\\
$\F$ & set of learned features: $\F=\{\F_0, \cdots, \F_h\}$\\
$\DepthRelFunc$ & maximum numbers of relational feature operators to use\\
\noalign{\vskip 5pt} 
\multicolumn{2}{l}{\bgcolored{\textbf{Notations for cPCA}}{0.73\linewidth}} \\
$\CovT$, $\CovB$ & covariance matrices\\
$\ContParam$ & contrast parameter\\ 
\bottomrule
\end{tabular}

%% file: 7_related_work.tex
\section{Related Work}
To the best of our knowledge, our work is the first to introduce the use of contrastive learning for networks and provide a general and interpretable method under this approach. There exists little work in the exact area. Thus, we here review typical NRL and contrastive learning techniques.

\subsection{Network Representation Learning (NRL)}
Various NRL methods have been developed for learning latent representations of network nodes and/or links. 
For a comprehensive description of NRL methods, refer to the recent survey papers, such as \citet{cai2018comprehensive} and \citet{zhang2018deep}.
Here we focus on describing the closely related work using inductive and cross-network embedding methods. 

\subsubsection{Inductive NRL}

GraphSAGE~\citep{hamilton2017inductive} is an inductive NRL method that shares many similar ideas with DeepGL~\citep{rossi2018deep}.
Analogous to the relational functions $\RelFunc$ in DeepGL, GraphSAGE learns \emph{aggregator functions}.
However, GraphSAGE proposes more complex aggregators using LSTM and max-pooling concepts, compared to DeepGL's simple aggregators (e.g., mean). 
Moreover, GraphSAGE tunes parameters required by the aggregators and matrices that decide the weight for each learned feature, instead of the feature pruning in DeepGL.
These differences might enable GraphSAGE to better capture complex characteristics of networks without manual parameter tuning; however, the learned features might be difficult to interpret.
FastGCN~\citep{chen2018fastgcn} takes a similar approach to GraphSAGE except that FastGCN employs node sampling to save memory space. 
Also, HetGNN~\citep{zhang2019heterogeneous} enhances the aggregators to learn representations of heterogeneous networks.
These methods, including other variants of graph neural networks~\citep{zhang2018deep} (e.g., GAT by \citealp{velivckovic2017graph}, h/cGAO by \citealp{gao2019graph}, and InfoGraph by \citealp{sun2020infograph}), still suffer from lack of interpretability in the learned features.
Although GNNExplainer~\citep{ying2019gnnexplainer} aims to provide interpretable explanations for predictions made by these methods, it does not support explaining the learned features themselves.

\subsubsection{Cross-Network Embedding}

The inductive methods learn the features that can be generalized for unobserved nodes or other networks from one input network. 
On the contrary, the cross-network methods generate embeddings directly from multiple input networks. 
Most of the cross-network methods focus on finding similarities of nodes across networks, such as for node classification~\citep{shen2019network}, network similarity calculation~\citep{ma2019deep}, and network alignment~\citep{heimann2018regal}.
While CrossMVA by \citet{chu2019cross} is developed mainly for network alignment, it can produce embeddings that contain both similarity and dissimilarity information. 
However, a major drawback of CrossMVA is that anchor nodes are necessary as inputs (i.e., at least we need to know a small portion of node-correspondence), which we cannot obtain in many cases (e.g., the example in \autoref{sec:analysis_example}).
Also, CrossMVA's embeddings of the dissimilarity information only preserve discriminative structures across networks; as a result, it cannot find unique patterns in a specific network.

\subsection{Contrastive Learning}
Unlike discriminant analysis, such as linear discriminant analysis~\citep{JiaNZ09}, which aims to discriminate samples based on their classes, contrastive learning~\citep{zou2013contrastive} focuses on finding salient patterns in one dataset compared to another.
Several machine learning methods have been extended for contrastive learning.
For example, there are contrastive versions of latent Dirichlet allocation~\citep{zou2013contrastive}, hidden Markov models~\citep{zou2013contrastive}, and regressions~\citep{ge2016rich}.
More recently, including cPCA~\citep{abid2018exploring}, contrastive learning methods for representation learning have been introduced~\citep{abid2018exploring,dirie2019contrastive,abid2019contrastive,severson2019unsupervised,fujiwara2020supporting,fujiwara2020contrastive,zhang2021visual,fujiwara2022interactive}.
For example, \citet{dirie2019contrastive} introduced contrastive multivariate singular spectrum analysis (cMSSA) for decomposition of time-series data.
Similar to cPCA, cMSSA could provide the interpretability by computing the PC loadings; however, cMSSA is not suitable for our case where we handle non-time series data. 
On the other hand, contrastive variational autoencoder (cVAE)~\citep{abid2019contrastive,severson2019unsupervised} can be used as a contrastive learning method in cNRL.
The strength of cVAE over cPCA is that it can find unique patterns in a target dataset even when its samples and latent features have nonlinear relationships.
However, cVAE relies on multiple layers of neural networks, and thus the results of cVAE are difficult to interpret as similar to other neural-network-based methods.
Therefore, to use cVAE for interpretable cNRL, we need additional effort to help interpret the results.

%% file: 5_1_case_studies.tex
\section{Experimental Evaluation}
In the previous sections, we have introduced the concepts of cNRL and i-cNRL, as well as the related work. 
We have also demonstrated the effectiveness of i-cNRL in comparing social networks in \autoref{sec:analysis_example}.
To further evaluate the method, we first test i-cNRL with synthetic datasets that are generated with popular network models.
Then, we demonstrate several analysis examples using i-cNRL with publicly available real-world datasets (see \autoref{table:network_info}). 
Lastly, we provide quantitative and qualitative comparisons among i-cNRL and other potential cNRL implementations.
In each subsection, we list only the information closely related to our findings. 
Throughout the evaluation, we focus on analyzing the first two cPCs in order to provide 2D embedding visualizations. Also, similar to PCA, the first two cPCs are usually more important than other cPCs.
Details of learning parameters and results are provided in \autoref{app:experiment_details}.

\input{5_0_evaluation_with_nw_models.tex}

\subsection{Case Studies}
\label{sec:case_studies}

\subsubsection{Case Study 1: Network Model Refinement}

Designing a network model that can simulate real-world networks is fundamental to understand network formation mechanisms, to perform hypothetical analyses (e.g., if there is a growth of the number of nodes, what will happen?), to generate more available datasets for machine learning, and more~\citep{goldenberg2010survey}. 
This case study demonstrates the usage of i-cNRL to guide a refinement of network models.

Here, we use a peer-to-peer (P2P) network, specifically the Gnutella peer-to-peer file sharing network~\citep{ripeanu2002mapping,leskovec2007graph} available in SNAP Datasets~\citep{snapnets}  (\texttt{\PtoPNwID} in \autoref{table:network_info}) as a modeling subject.
Once we have a P2P network generation model, we can use it for analyzing network robustness, studying effective searching strategies on a P2P network, etc~\citep{liu2009efficient}.

\begin{figure}[tb]
	\centering
	\captionsetup[subfloat]{width=0.4\linewidth}
    \subfloat[$\GraphT$: p2p-Gnutella08, $\GraphB$: Price 2]{
        \includegraphics[width=0.4\linewidth]{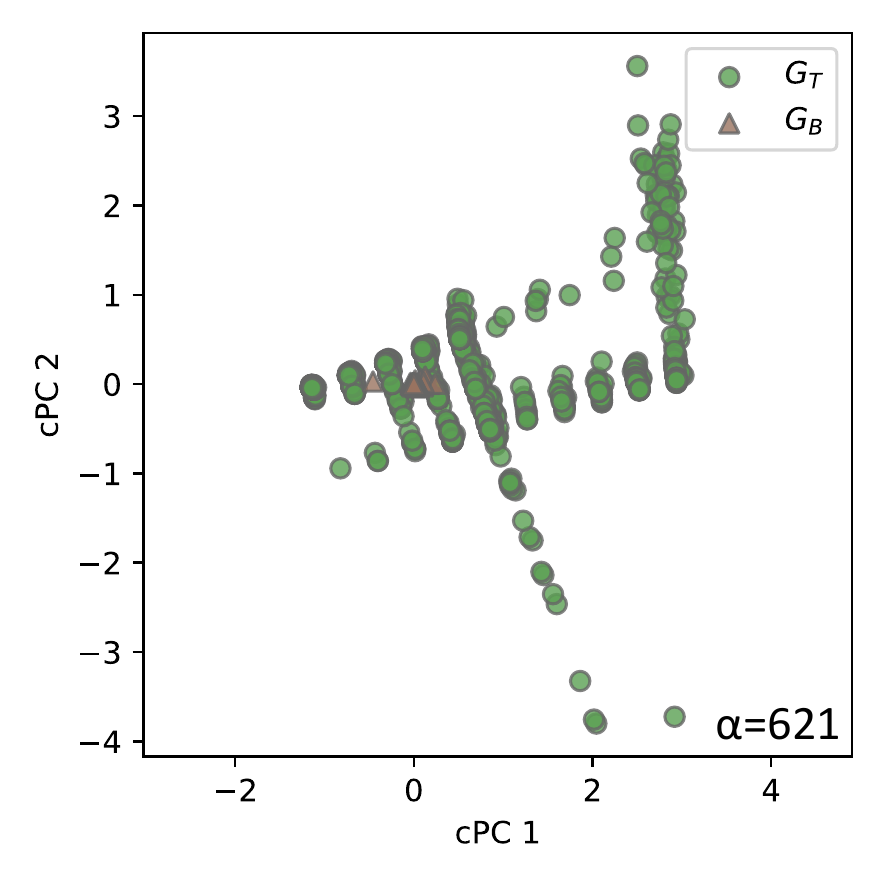}
        \label{fig:p2p_price_cpca}
    }
    \subfloat[$\GraphT$: p2p-Gnutella08, $\GraphB$: Price 2]{
        \includegraphics[width=0.45\linewidth]{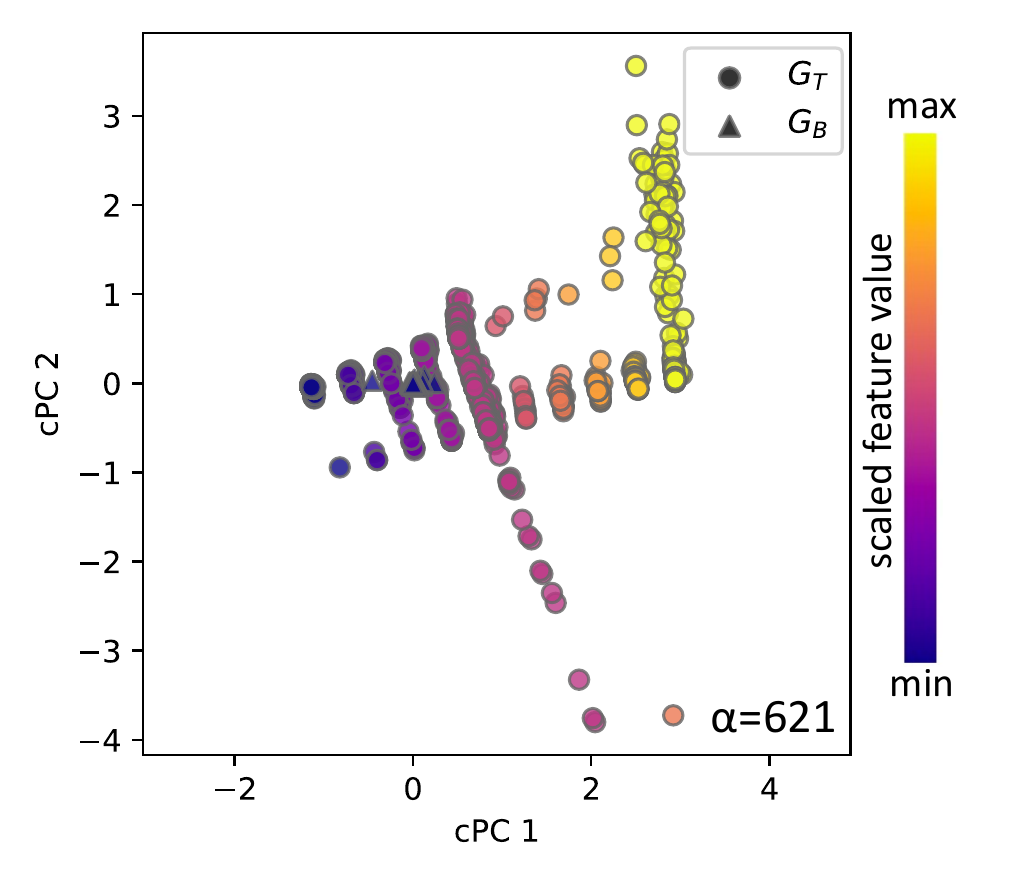}
        \label{fig:p2p_price_cpca_colored}
    }
    \caption{Results for Case Study 1. (a) presents the 2D embedding by i-cNRL. (b) shows the nodes in (a) colored by the $k$-core number (\texttt{\PtoPVsPriceTwo-1} in \autoref{table:top_loadings}).}
	\label{fig:p2p_vs_price}
\end{figure}

P2P networks are often scale-free~\citep{liu2009efficient}, so we use the Price's model~\citep{newman2018networks} to mimic a P2P network. 
To identify the characteristics that the Price's model does not simulate well, we set the P2P network (\texttt{\PtoPNwID}) as $\GraphT$ and the Price network (\texttt{\PriceTwoNwID}) as $\GraphB$.

The result is shown in \autoref{fig:p2p_vs_price}\subref{fig:p2p_price_cpca}.
From the cPC loadings in \autoref{table:top_loadings}, we notice that the $k$-core number (\texttt{\PtoPVsPriceTwo-1}) has a strong contribution to \texttt{cPC~1}. 
Thus, we colorcode the result based on the $k$-core number, as shown in \autoref{fig:p2p_vs_price}\subref{fig:p2p_price_cpca_colored}.
We can clearly see that the P2P network has variations in the $k$-core number, but the Price network does not.
Because the $k$-core number indicates that a node at least connects to other $k$ nodes, the Price network makes a significant difference in the network robustness from the P2P network.

From the result above, we decide to refine the Price model to generate various $k$-core numbers. 
As discussed in \autoref{sec:test_models}, the problem comes from the fact that the Price's model always adds a new node with a fixed number of links.
Similar to the dual-Barab\'{a}si-Albert model by \citet{moshiri2018dual}, we can avoid the problem by attaching a new node to a variable number of links according to a probability distribution. 
Specifically, we design an enhanced version of Price's model to select the number of links from 1 to 10 with specified probabilities (for details, refer to \autoref{sec:net_gen_model_prameters}).
Then, we generate a network with this model, which is referred to as the Enhanced Price (\texttt{\EPriceNwID}) network in \autoref{table:network_info}. 
Next, we apply i-cNRL to the P2P (as $\GraphT$) and Enhanced Price (as $\GraphB$) networks. 
The resultant cPC loadings are listed in \autoref{table:top_loadings}.
While $\GraphT$ seems to still have the uniqueness in degree centralities, it does not in the $k$-core number. 
By iteratively performing refinement procedures such as the one above, we can build a better network model to simulate real-world networks.

\subsubsection{Case Study 2: Comparison of Two Networks} 

In this case study, we compare ``interactome'' networks---networks of physical DNA\mbox{-}, RNA-, and protein-protein interactions~\citep{yu2008high}.
Specifically, we compare two interactome networks, Combined-AP/MS (\texttt{\CombinedAPMSNwID} in \autoref{table:network_info}) and LC-multiple (\texttt{\LCMultiNwID}), available in CCSB Interactome Database~\citep{ccsb}.
Both networks represent the interactome of the yeast \textit{S.~cerevisiae}; however, they are obtained through different analysis approaches. 
Combined-AP/MS is generated from two studies using a ``high-throughput'' approach, specifically, affinity purification/mass spectrometry (AP/MS)~\citep{collins2007toward}. 
In contrast, LC-multiple is the literature-curated (LC) network from multiple ``low-throughput'' experiments~\citep{yu2008high,reguly2006comprehensive}.  
Because each analysis approach has its own strength in identifying the yeast's interactions, the generated networks may vary~\citep{yu2008high}. 
Comparing these networks is essential to understand the quality and characteristics of each approach~\citep{yu2008high}. 

\begin{figure}[tb]
	\centering
	\captionsetup[subfloat]{width=0.3\linewidth}
    \subfloat[$\GraphT$: LC-multiple,\newline $\GraphB$: Combined-AP/MS]{
        \includegraphics[width=0.3\linewidth]{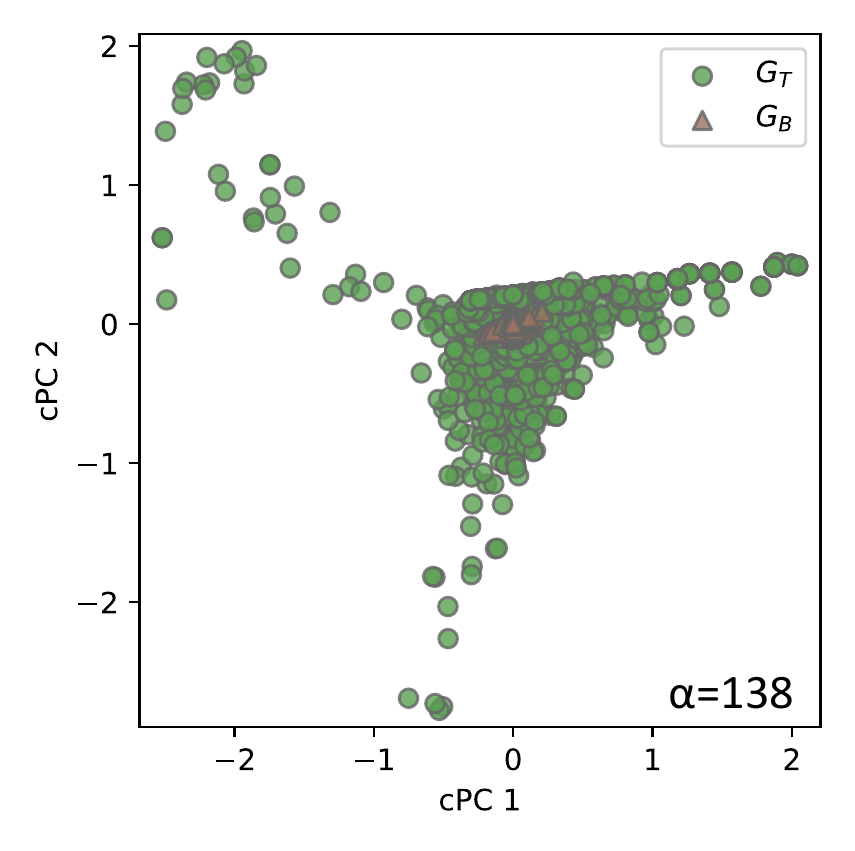}
        \label{fig:lc_collins_cpca}
    }
    \subfloat[$\GraphT$: LC-multiple,\newline $\GraphB$: Combined-AP/MS]{
        \includegraphics[width=0.3\linewidth]{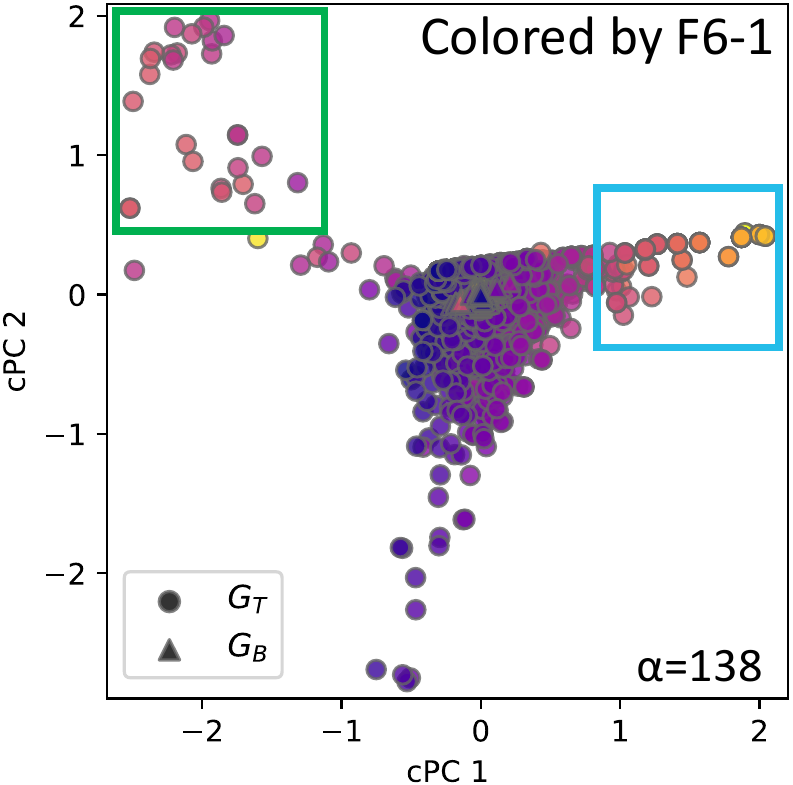}
        \label{fig:lc_collins_cpca_colored}
    }
    \subfloat[$\GraphT$: LC-multiple,\newline $\GraphB$: Combined-AP/MS]{
        \includegraphics[width=0.34\linewidth]{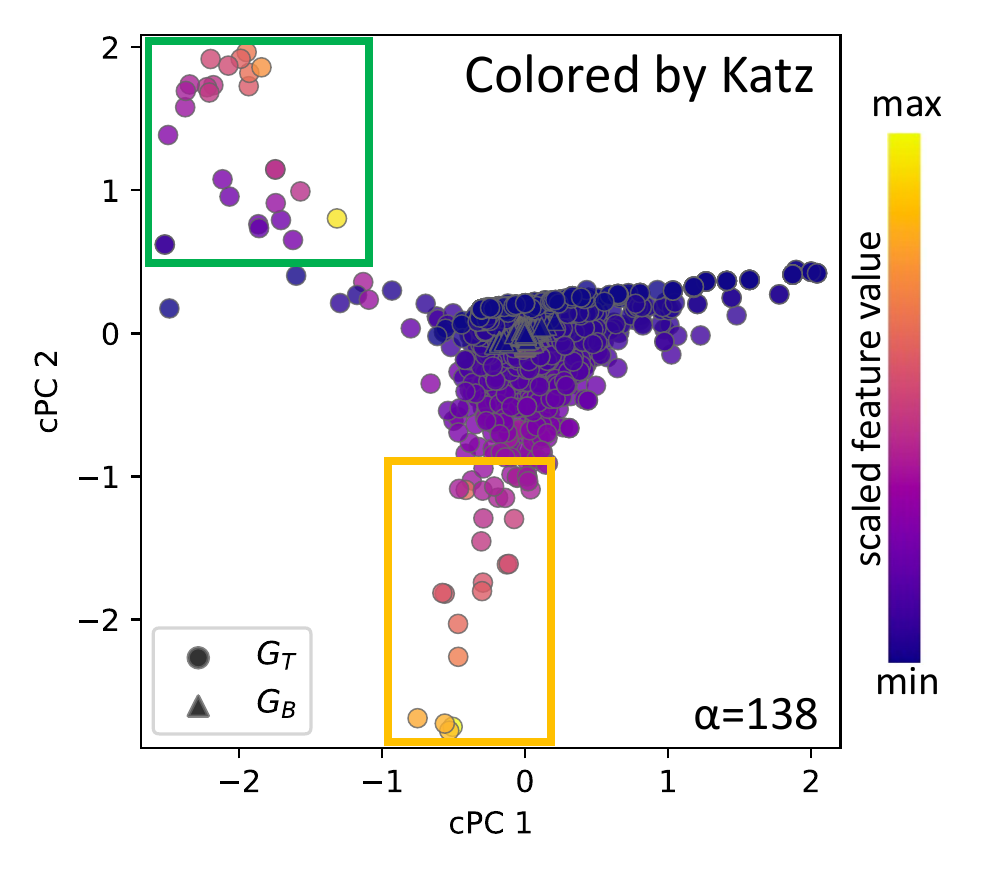}
        \label{fig:lc_collins_cpca_colored2}
    }
    \\
    \subfloat[$\GraphT$: LC-multiple]{
        \includegraphics[width=0.49\linewidth]{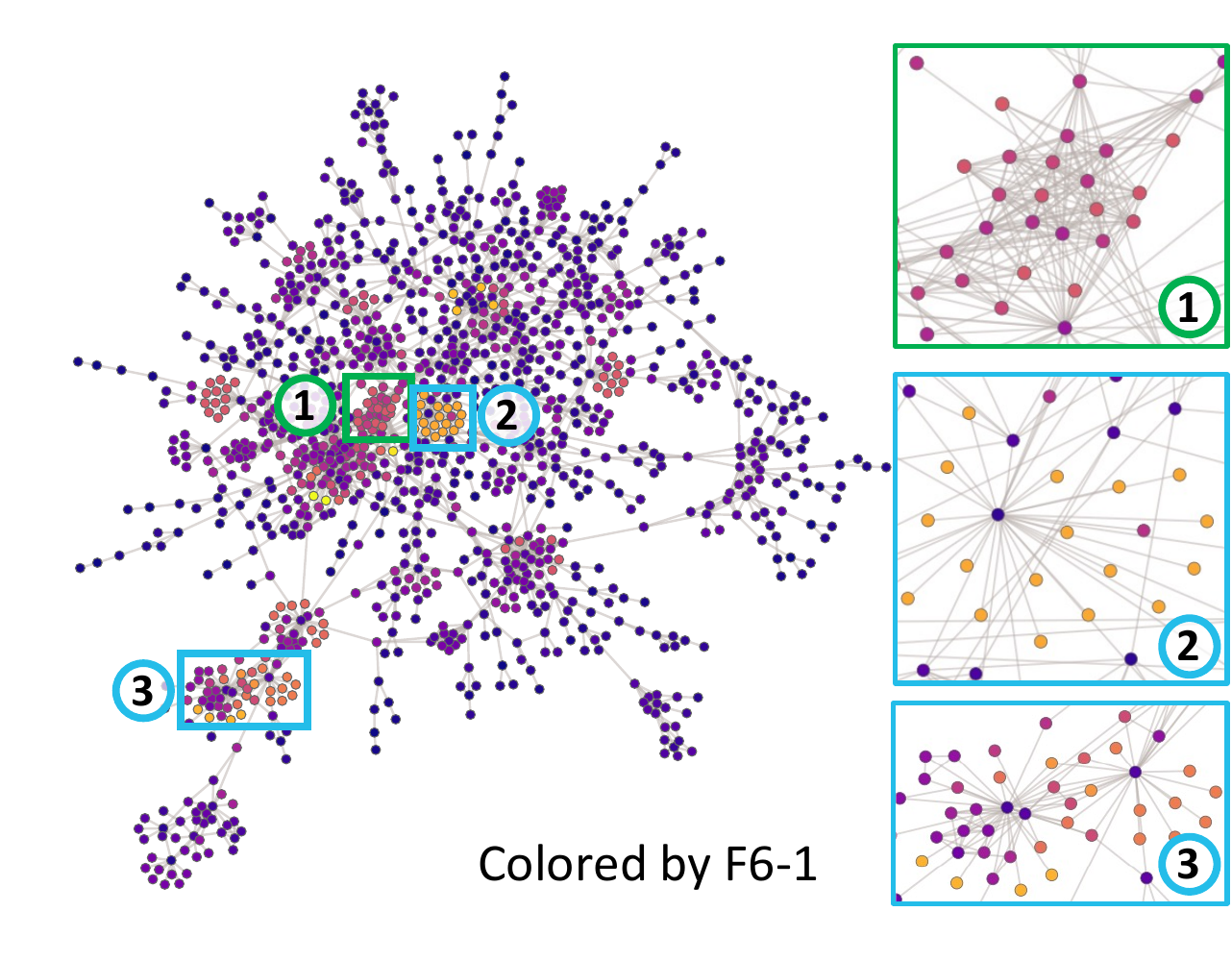}
        \label{fig:lc_colored}
    }
    \hspace{10pt}
    \subfloat[$\GraphB$: Combined-AP/MS]{
        \includegraphics[width=0.42\linewidth]{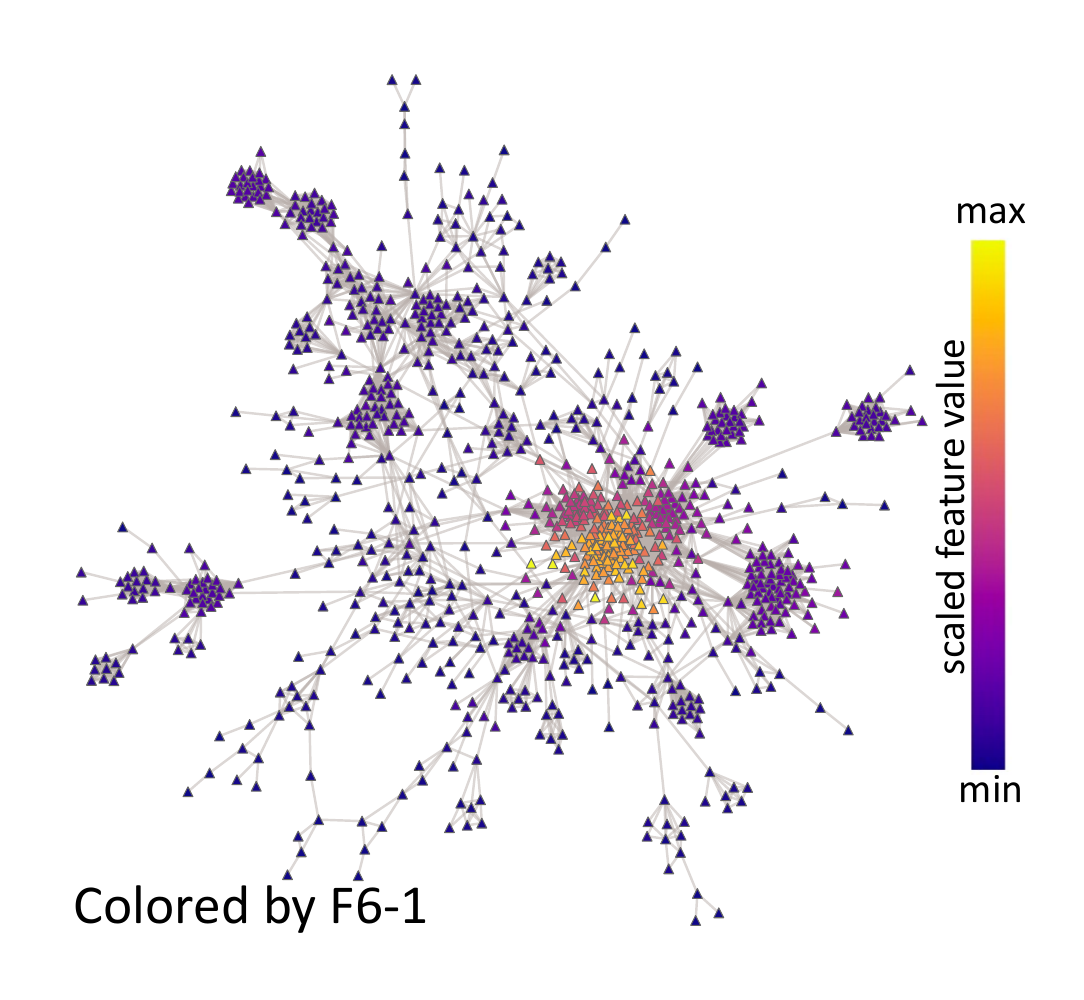}
        \label{fig:collins_colored}
    }
    \\
    \subfloat[$\GraphT$: LC-multiple]{
        \includegraphics[width=0.49\linewidth]{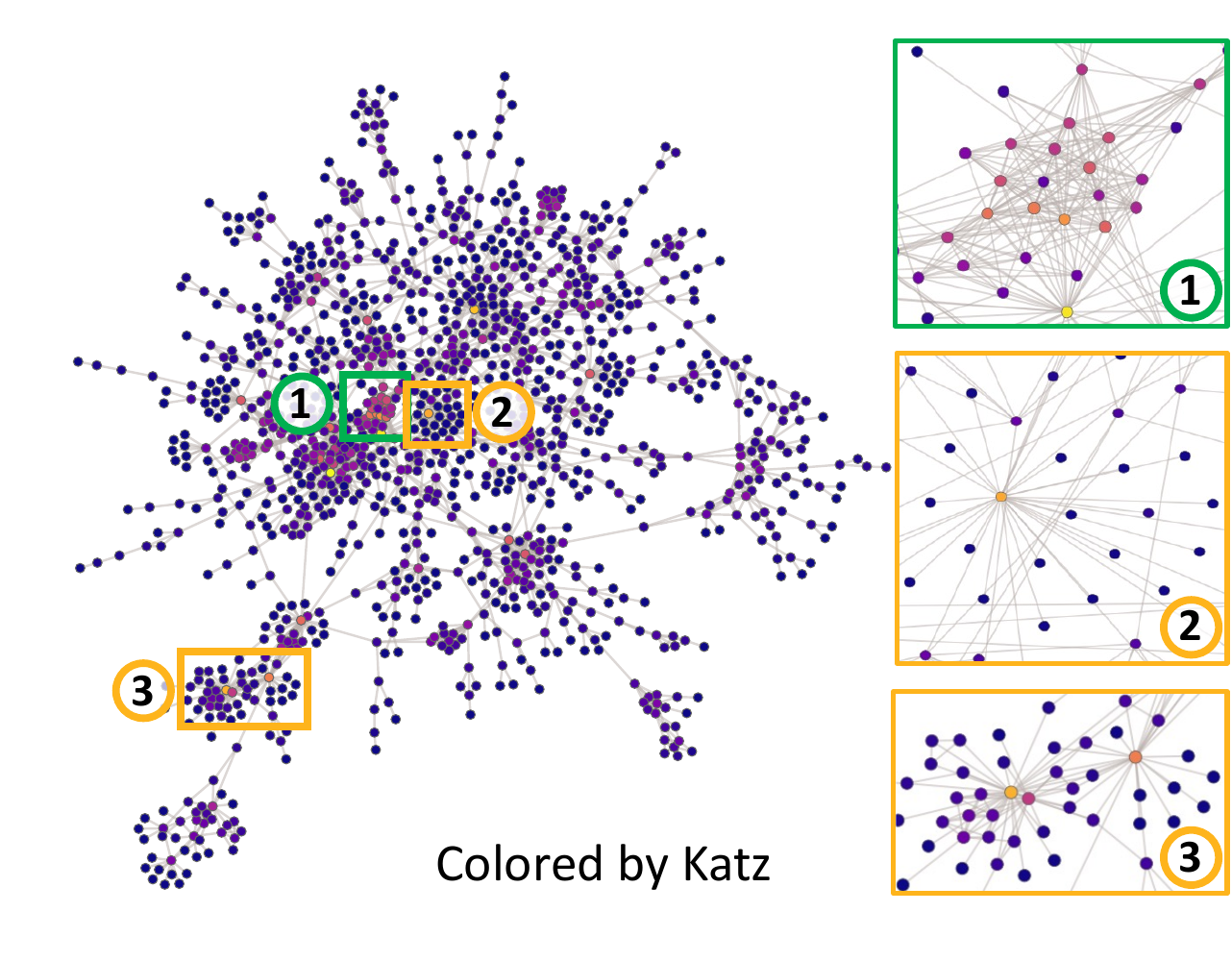}
        \label{fig:lc_colored2}
    }
    \hspace{10pt}
    \subfloat[$\GraphB$: Combined-AP/MS]{
        \includegraphics[width=0.42\linewidth]{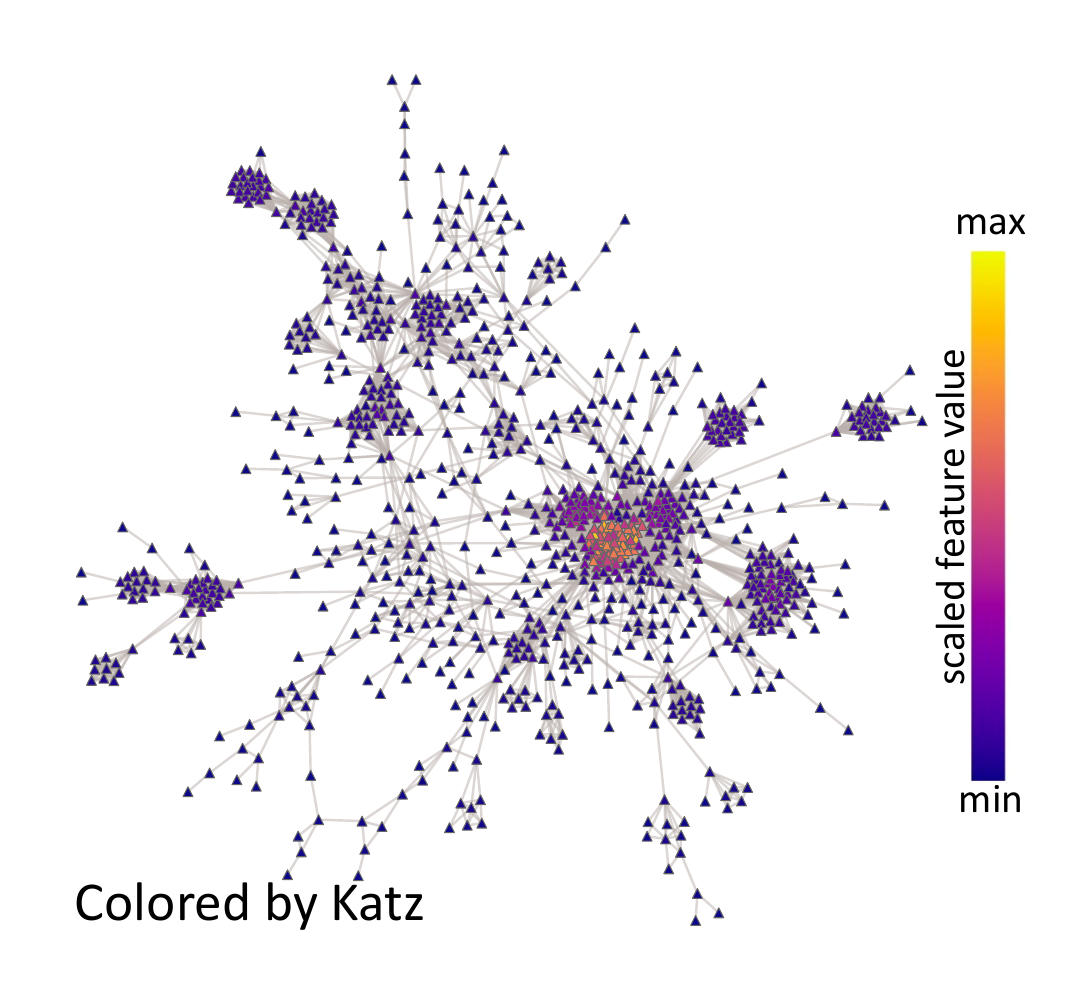}
        \label{fig:collins_colored2}
    }
    \caption{Results for Case Study 2. (a) presents the 2D embedding by i-cNRL. (b) shows the nodes in (a) colored by the feature---$\RelFunc$: $(\RelFeatOpe{}{\Mean})(\BaseFeat)$, $\BaseFeat$: the Katz centrality (\texttt{\LCMultiVsCombinedAPMS-1} in \autoref{table:top_loadings}). 
    (c) is colored by the Katz centrality.
    (d--g) show the network structures with the same colorcoding (d and e: \texttt{\LCMultiVsCombinedAPMS-1}, f and g: Katz).
    }
	\label{fig:lc_vs_collins}
\end{figure}

Here we analyze the uniqueness in LC-multiple by using LC-multiple and Combined-AP/MS as $\GraphT$ and $\GraphB$, respectively.
The 2D embedding result by i-cNRL is shown in \autoref{fig:lc_vs_collins}\subref{fig:lc_collins_cpca}. 
We first notice that, in $\GraphT$, there are two distinct regions: one spreading out towards the top-left and the other in the bottom-right quadrant.
To understand why this pattern appears, we obtain the cPC loadings (\autoref{table:top_loadings}) and colorcode the nodes based on values of the feature that has the top cPC loading for \texttt{cPC~1} (i.e., \texttt{\LCMultiVsCombinedAPMS-1}, $\RelFunc$: $(\RelFeatOpe{}{\rm mean})(\BaseFeat)$ and $\BaseFeat$: the Katz centrality).
The result is shown in \autoref{fig:lc_vs_collins}\subref{fig:lc_collins_cpca_colored}. 
We observe that either going to the left or right side along \texttt{cPC~1} tends to produce a high value of this feature, as annotated with the green and teal rectangles, respectively.
While this feature has a strong positive loading for \texttt{cPC~1}, another feature in \autoref{table:top_loadings}---\texttt{\LCMultiVsCombinedAPMS-2}, $\RelFunc$: $(\RelFeatOpe{}{\rm mean})(\BaseFeat)$ and $\BaseFeat$: the eigenvector centrality---has a strong negative loading.
Therefore, if a node has a higher value for \texttt{\LCMultiVsCombinedAPMS-2}, it tends to be placed on the more left side in \autoref{fig:lc_vs_collins}\subref{fig:lc_collins_cpca_colored}.
This indicates that the green rectangle region in \autoref{fig:lc_vs_collins}\subref{fig:lc_collins_cpca_colored} seems to have high values for both of these features while the teal region has low values for the latter feature (\texttt{\LCMultiVsCombinedAPMS-2}). 
This could happen because the eigenvector centrality tends to be low when a node is in a weakly connected region while the Katz centrality is high whenever a node is linked by many others~\citep{newman2018networks}.

To visually observe the above patterns, we draw the network structures of $\GraphT$ and $\GraphB$ with scalable force directed placement~\citep{hu2005efficient} and then color them based on the values of \texttt{\LCMultiVsCombinedAPMS-1}, as shown in \autoref{fig:lc_vs_collins}\subref{fig:lc_colored} and \subref{fig:collins_colored}. 
We here only show the largest component of each network (i.e., the nodes connected with only several nodes are filtered out).
\autoref{fig:lc_vs_collins}\subref{fig:collins_colored} shows that one strongly connected region around the center contains all nodes with high feature values. 
On the other hand, in \autoref{fig:lc_vs_collins}\subref{fig:lc_colored}, multiple regions contain nodes with high feature values. 
To further investigate this pattern, we select the nodes corresponding to the green and teal regions in \autoref{fig:lc_vs_collins}\subref{fig:lc_collins_cpca_colored} and then highlight these nodes in \autoref{fig:lc_vs_collins}\subref{fig:lc_colored}.
Afterward, we zoom into the related regions of the highlighted nodes.
\autoref{fig:lc_vs_collins}\subref{fig:lc_colored}-\circled{1} shows a region related to the nodes in the green rectangle, while \autoref{fig:lc_vs_collins}\subref{fig:lc_colored}-\circled{2} and \circled{3} are two example regions related to the teal rectangle region. 
We can see that the nodes in \autoref{fig:lc_vs_collins}\subref{fig:lc_colored}-\circled{1} are strongly connected, but not in \autoref{fig:lc_vs_collins}\subref{fig:lc_colored}-\circled{2} and \circled{3}. 
From these observations, i-cNRL reveals that only $\GraphT$ has two different types of nodes linked to the high Katz centrality node(s) in either strongly or weakly connected region.

\begin{table}[tb]
    \centering
    \footnotesize
    \caption{The features with the top-3 absolute loadings for \texttt{cPC~2} for Case Study 2.}
    \label{table:top_loadings_pc2}
    \input{tables/top_loadings_pc2}
\end{table}

Similarly, we further interpret \texttt{cPC~2}. 
As shown in \autoref{table:top_loadings_pc2}, the base features of the Katz, total-degree, and eigenvector centralities strongly contribute to \texttt{cPC~2}.
From the 2D embedding result in \autoref{fig:lc_vs_collins}\subref{fig:lc_collins_cpca_colored2}, which is colorcoded based on the Katz centrality, we can see that the Katz centrality also shows high values in two regions, as annotated with the green and orange rectangles. 
As with the analysis of \texttt{cPC~1}, the Katz and eigenvector centralities have strong negative and positive loadings to \texttt{cPC~2}, respectively.
To investigate the two regions, we generate visualizations corresponding to the analysis of \texttt{cPC~1} (\autoref{fig:lc_vs_collins}\subref{fig:lc_colored2} and \subref{fig:collins_colored2}). 
From these visualizations, we can see that, in the orange rectangle area of \autoref{fig:lc_vs_collins}\subref{fig:lc_collins_cpca_colored2}, \texttt{cPC~2} seems to capture central nodes in the aforementioned weekly connected region.
In summary, i-cNRL highlights the three different types of nodes unique in the target network---(1) nodes in the strongly connected regions, (2) central nodes, and (3) surrounding nodes in the weakly connected regions.  

\begin{figure}[tb]
	\centering
    \subfloat[$\GraphT$: Day 2]{
        \includegraphics[width=0.28\linewidth]{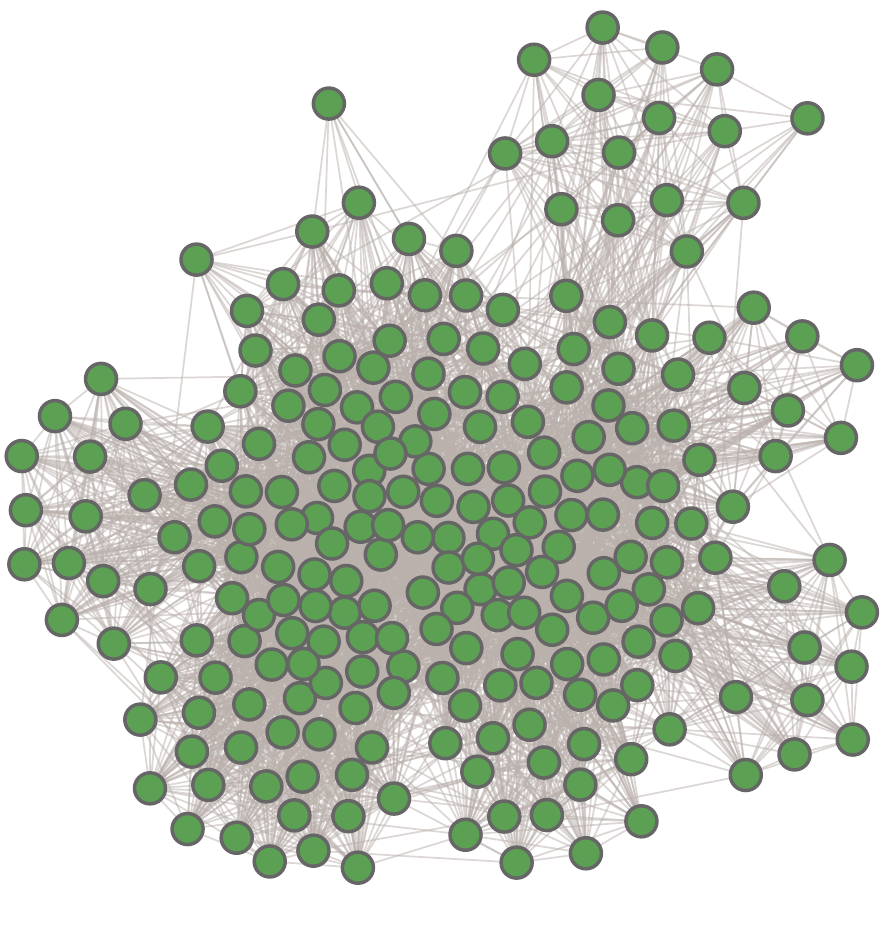}
        \label{fig:day2}
    }
    \subfloat[$\GraphB$: Day 1]{
        \includegraphics[width=0.28\linewidth]{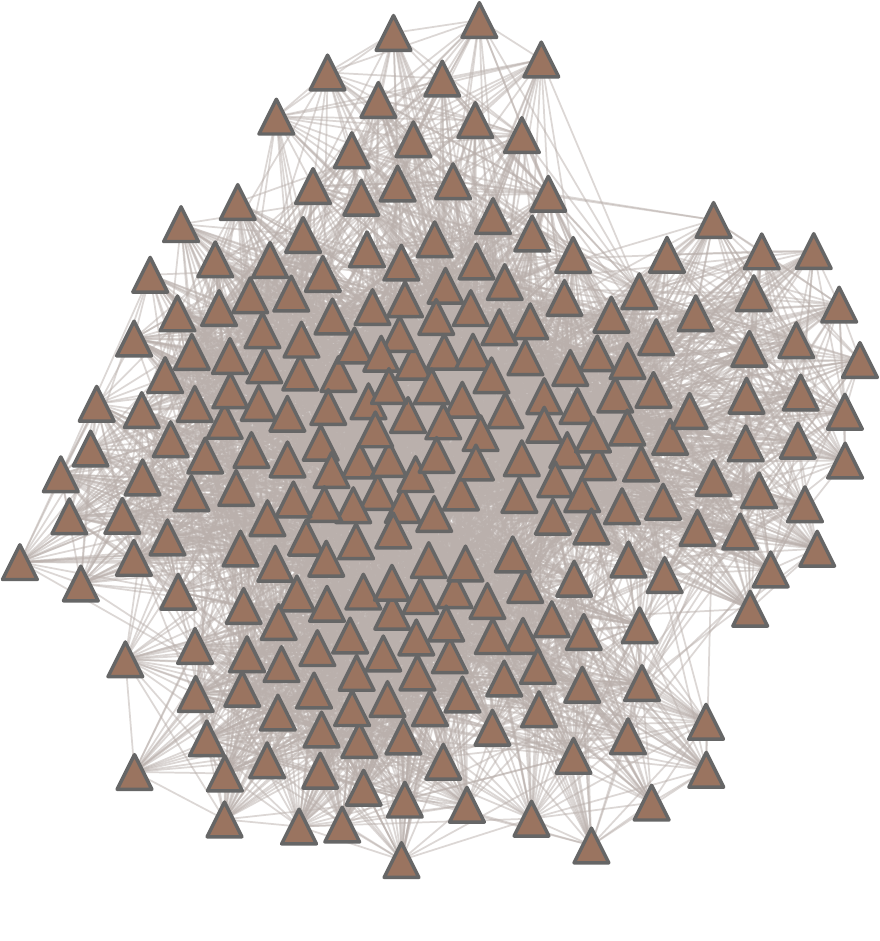}
        \label{fig:day1}
    }
    \subfloat[$\GraphT$: Day 2, $\GraphB$: Day 1]{
        \includegraphics[width=0.38\linewidth]{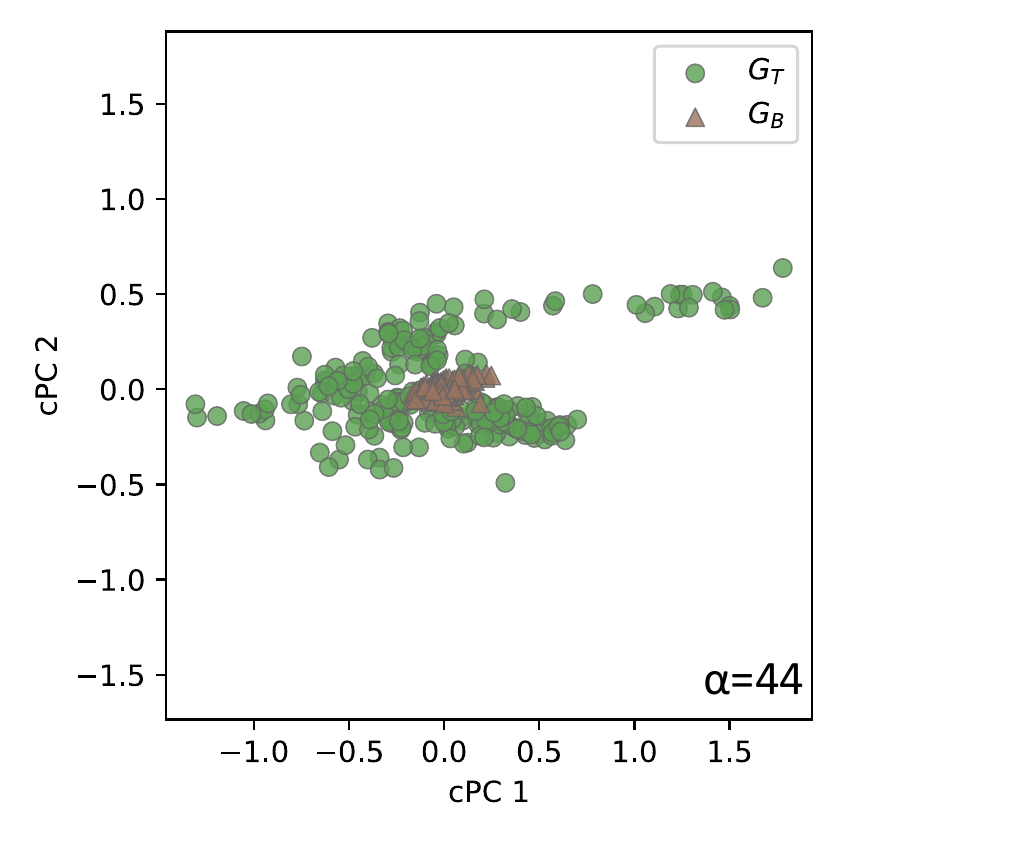}
        \label{fig:day2_day1_cpca}
    }
    \\
    \subfloat[$\GraphT$: Day 2]{
        \includegraphics[width=0.28\linewidth]{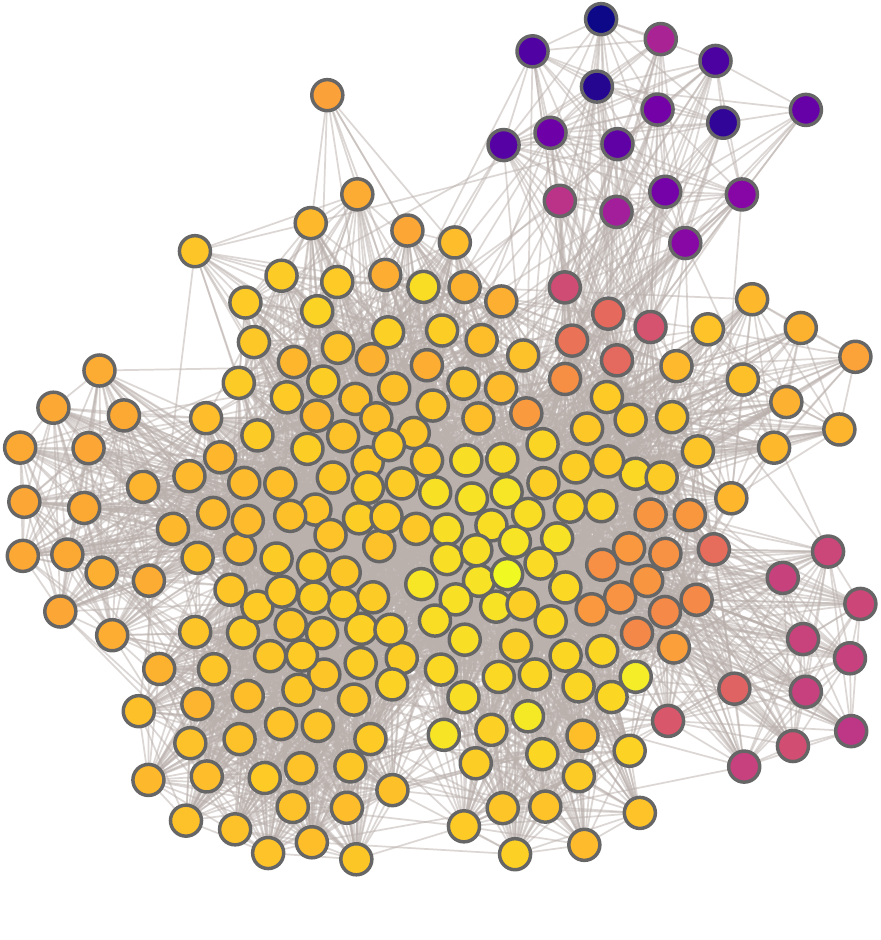}
        \label{fig:day2_colored}
    }
    \subfloat[$\GraphB$: Day 1]{
        \includegraphics[width=0.28\linewidth]{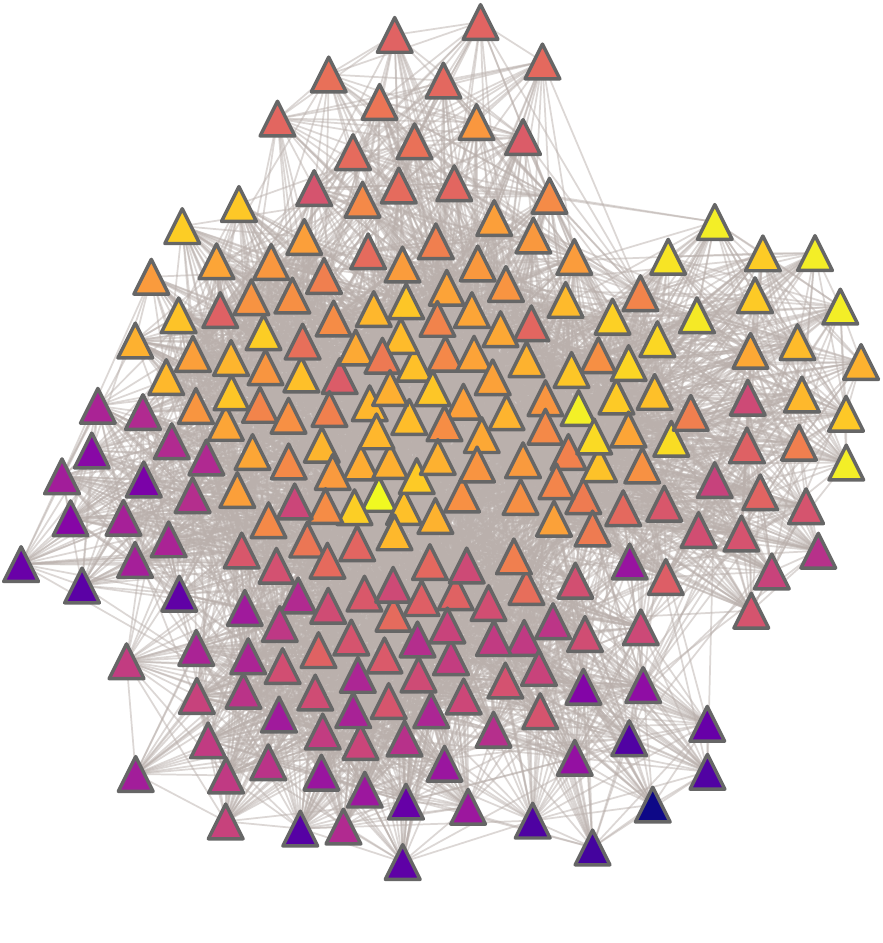}
        \label{fig:day1_colored}
    }
    \subfloat[$\GraphT$: Day 2, $\GraphB$: Day 1]{
        \includegraphics[width=0.38\linewidth]{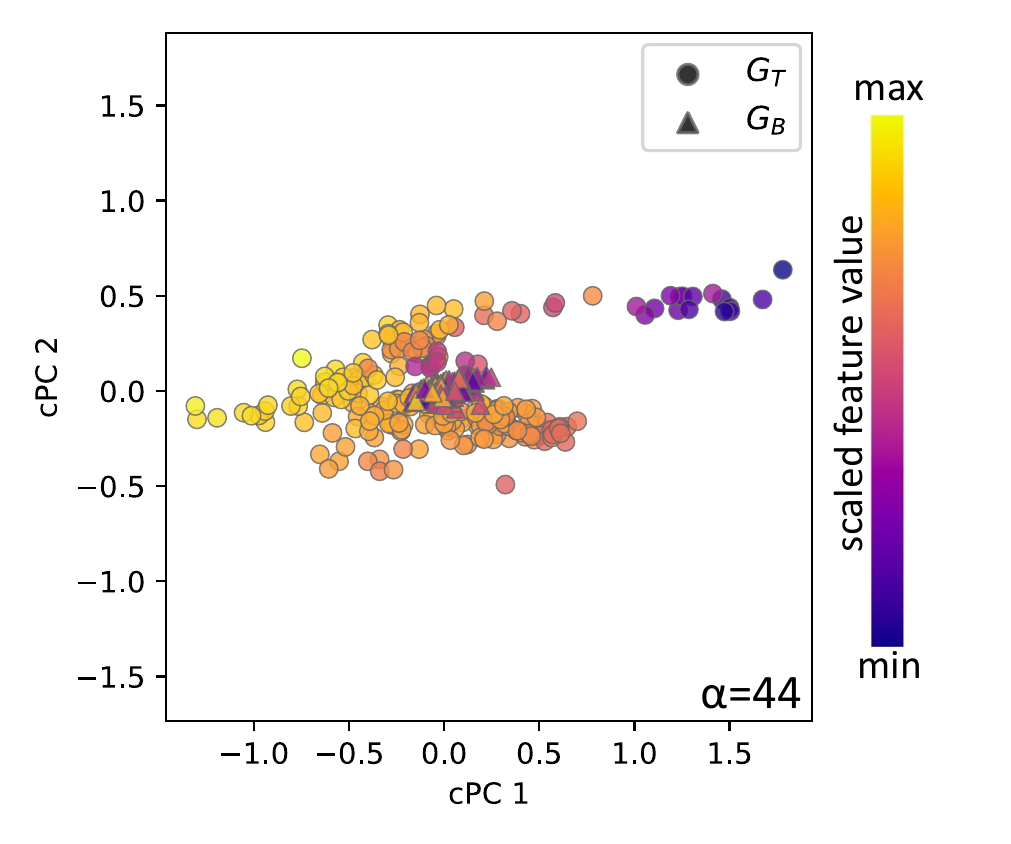}
        \label{fig:day2_day1_cpca_colored}
    }
    \\
    \subfloat[$\GraphT$: Day 2]{
        \includegraphics[width=0.28\linewidth]{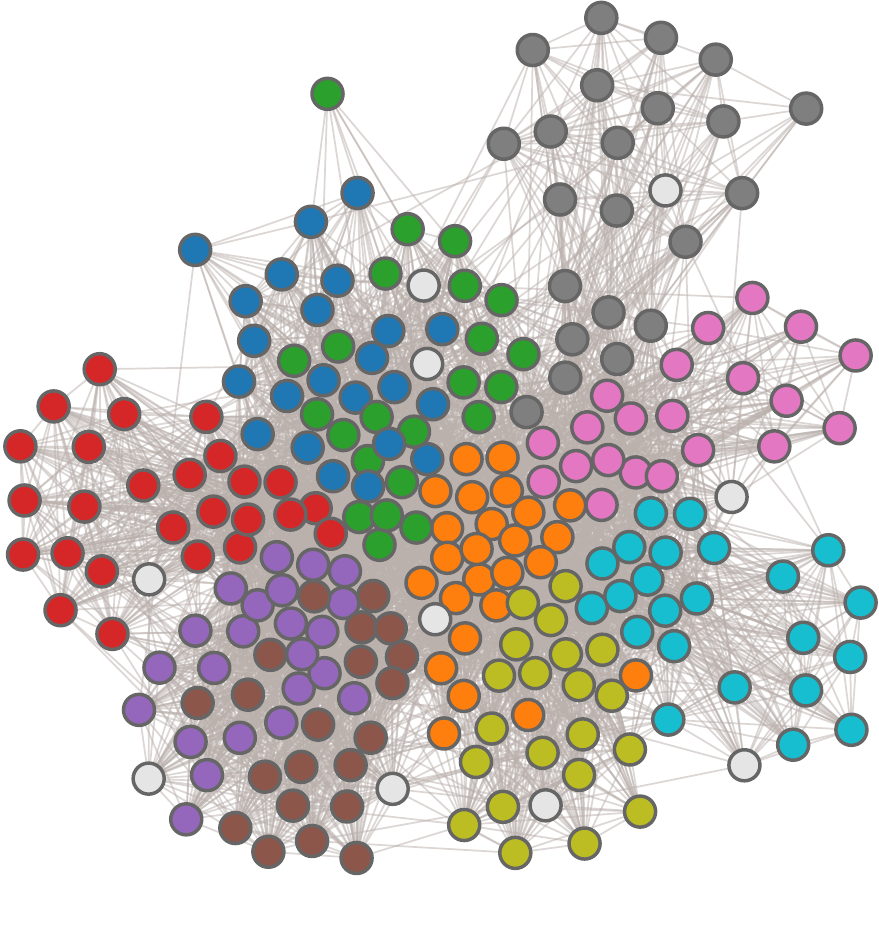}
        \label{fig:day2_class}
    }
    \subfloat[$\GraphB$: Day 1]{
        \includegraphics[width=0.28\linewidth]{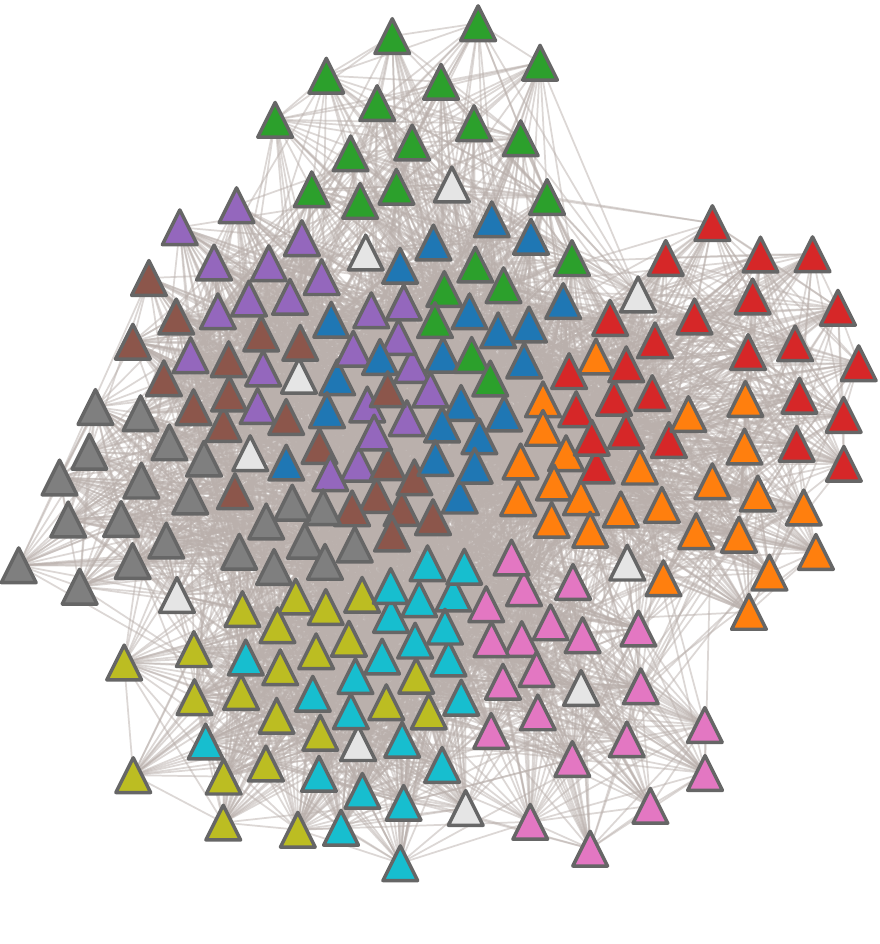}
        \label{fig:day1_class}
    }
    \subfloat[$\GraphT$: Day 2, $\GraphB$: Day 1]{
        \includegraphics[width=0.38\linewidth]{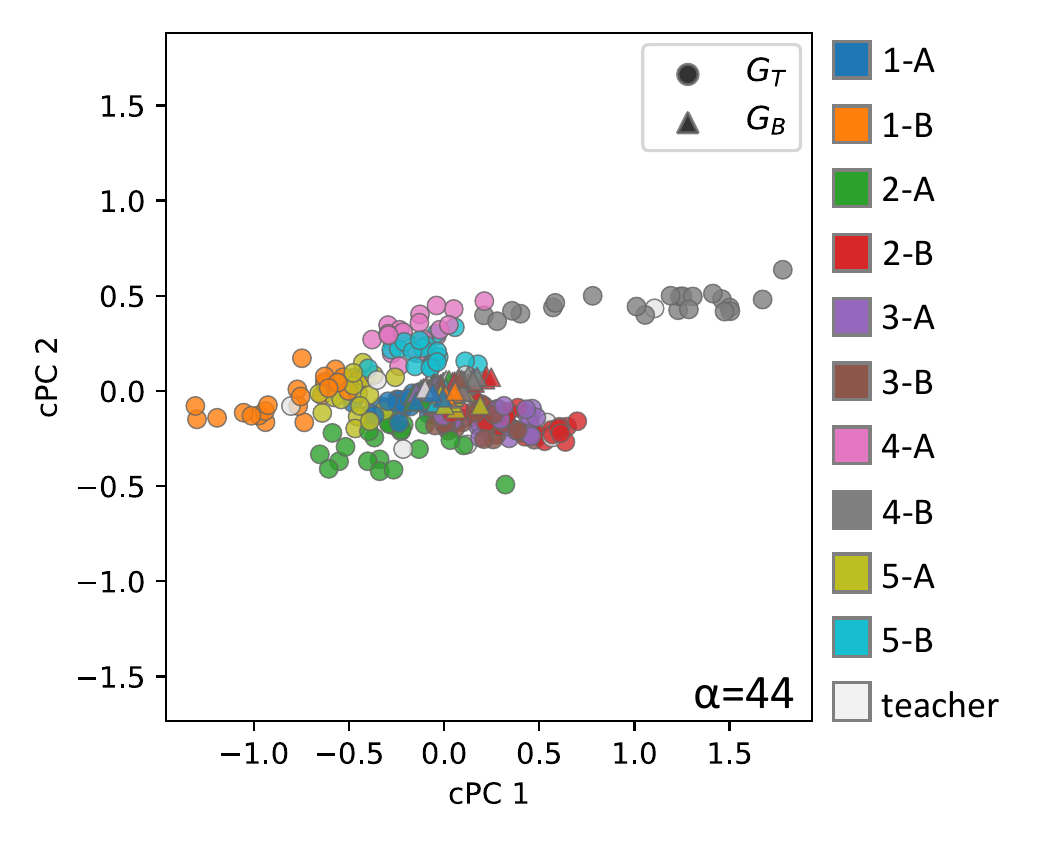}
        \label{fig:day2_day1_cpca_class}
    }
    \caption{Results for Case Study 3. (a) and (b) show the network structures of $\GraphT$ and $\GraphB$. (c) presents the 2D embedding by i-cNRL. 
    (d-f) show the colorcoded nodes in (a-c) based on the feature---$\RelFunc$: $(\RelFeatOpe{}{\Mean} \circ \RelFeatOpe{}{\Max})(\BaseFeat)$, $\BaseFeat$: the PageRank (\texttt{\SchSecondVsFirst-1} in \autoref{table:top_loadings}). (g-i) show the nodes colored by the class name where the first number indicates the grade (e.g., `1-A' is the first grade class). The networks include `teacher' nodes.}
	\label{fig:school}
\end{figure}

\subsubsection{Case Study 3: Analysis of Network Changes}

As an example of analyzing dynamic networks, we compare two different days of contact networks in a primary school~\citep{stehle2011high,sociopatterns}.
The networks represent face-to-face contact patterns between students and teachers, which are collected with RFID devices. 
Information of the network at each day is listed in \autoref{table:network_info} (\texttt{\SchFirstDayNwID} and \texttt{\SchSecondDayNwID}). 
\autoref{fig:school}\subref{fig:day2} and \subref{fig:day1} visualize the network structures drawn with scalable force directed placement.
Also, these networks have multiple node attributes including genders, grades, and class names. 
In addition to multiple network centralities, we utilize the attribute information by including gender as the base feature, i.e., encoding `male', `female', and `unknown' as -1, 1, and 0, respectively.

To analyze changes in contact patterns, we set the networks of the second day and the first day as $\GraphT$ and $\GraphB$, respectively.
\autoref{fig:school}\subref{fig:day2_day1_cpca} shows the 2D embedding result.
To interpret $\GraphT$'s unique patterns, we review the cPC loadings listed in \autoref{table:top_loadings} and colorcode the nodes in \autoref{fig:school}\subref{fig:day2}, \subref{fig:day1}, and \subref{fig:day2_day1_cpca} based on the learned feature \texttt{\SchSecondVsFirst-1}---$\RelFunc$: $(\RelFeatOpe{}{\rm mean} \circ \RelFeatOpe{}{\rm max})(\BaseFeat)$, $\BaseFeat$: PageRank.
The results are shown in \autoref{fig:school}\subref{fig:day2_colored}, \subref{fig:day1_colored}, and \subref{fig:day2_day1_cpca_colored}.
We can see that i-cNRL discovers that $\GraphT$ has both strongly (colored with more yellow in \autoref{fig:school}\subref{fig:day2_colored} and \subref{fig:day2_day1_cpca_colored}) and weakly connected regions from others (colored with more purple), while all of $\GraphB$'s nodes have relatively strong connections between each other, as seen in the laid-out result in \autoref{fig:school}\subref{fig:day1}. 

According to the study by \citet{stehle2011high}, the students tended to have more contact within the same class than between classes. 
To relate the class information and the found unique patterns, we colorcode the nodes (i.e., students) based on their class, as shown in \autoref{fig:school}\subref{fig:day2_class}, \subref{fig:day1_class}, and \subref{fig:day2_day1_cpca_class}.
From these results, we notice that i-cNRL well separates groups of students who have less (e.g., gray, pink, or teal nodes) and more (e.g., orange nodes) contact between classes in $\GraphT$.

%% file: 5_0_evaluation_with_nw_models.tex
\begin{figure}[tb]
	\centering
    \subfloat[$\GraphT$: Price, $\GraphB$: Random]{
        \includegraphics[width=0.4\linewidth]{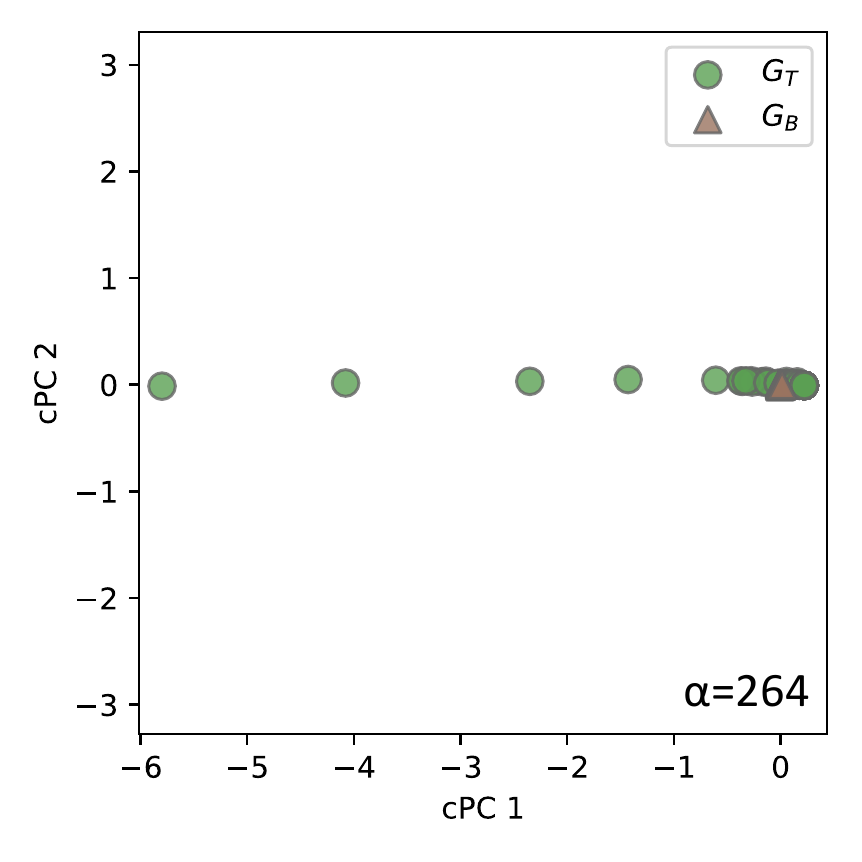}
        \label{fig:ba_rand_cpca}
    }
    \subfloat[$\GraphT$: Random, $\GraphB$: Price]{
        \includegraphics[width=0.4\linewidth]{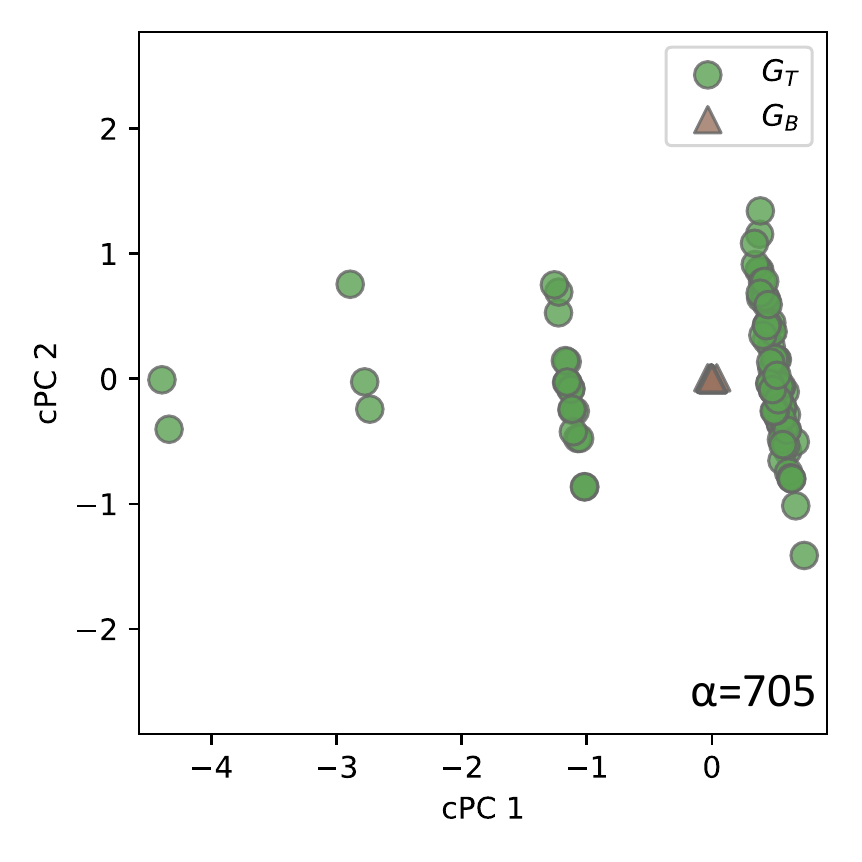}
        \label{fig:rand_ba_cpca}
    }
    \caption{Results for \autoref{sec:test_models} with 2D embeddings by i-cNRL.}
	\label{fig:network_models}
\end{figure}

\subsection{Evaluation with Network Models}
\label{sec:test_models}

We apply i-cNRL to compare two types of synthetic networks: \emph{random} and \emph{scale-free} networks (\texttt{\RandomNwID} and \texttt{\PriceNwID} in \autoref{table:network_info}). 
We generate the random and scale-free networks with the Gilbert's random graph~\citep{barabasi2016network} and the Price's preferential attachment models~\citep{newman2018networks}, respectively. 
We produce two 2D embedding results, using one network as $\GraphT$ and the other as $\GraphB$, as shown in \autoref{fig:network_models}\subref{fig:ba_rand_cpca} and \subref{fig:rand_ba_cpca}. 
Each of the results shows unique patterns in $\GraphT$. 
The cPC loadings in \autoref{table:top_loadings} show that the Price network's unique patterns are related to the degree centralities (e.g., total-degree). 
This seems to be due to the fact that most nodes have the same number of links in a random network while a scale-free network contains hubs with a large number of links.
In contrast, we can see that the random network's uniqueness is mostly related to $k$-core numbers. 
This is because the Price's model generates a network by adding a new node and then connecting it to other fixed number of nodes (e.g., 3 nodes) which are selected with a certain computed probability. 
As a result, all nodes in the network have the same $k$-core numbers (e.g., 3-core).

\begin{table}[tb]
    \centering
    \footnotesize
    \caption{The features with the top-3 absolute loadings for \texttt{cPC~1} for different pairs of networks highlighted in gray.}
    \label{table:top_loadings}
    \input{tables/top_loadings.tex}
\end{table}

%% file: tables/top_loadings.tex
\begin{tabular}{rllrr}
\toprule
ID & relational function $\RelFunc$ & base feature $\BaseFeat$ & \texttt{cPC 1} & \texttt{cPC 2} \\
\midrule

\multicolumn{5}{l}{
\bgcolored{$\GraphT$: Price, $\GraphB$: Random (\autoref{sec:test_models})}{0.62\linewidth}
} \\
\newline
\texttt{\PriceVsRandom-1} & $(\BaseFeat)$ & total-degree & -0.79 & -0.01 \\
\texttt{\PriceVsRandom-1} & $(\BaseFeat)$ & out-degree & 0.52 & 0.00 \\
\texttt{\PriceVsRandom-3} & $(\BaseFeat)$ & Katz & 0.24 & -0.71 \\
\noalign{\vskip 4pt} 
\multicolumn{5}{l}{
\bgcolored{
$\GraphT$: Random, $\GraphB$: Price (\autoref{sec:test_models})}{0.62\linewidth}
} \\ 
\texttt{\RandomVsPrice-1} & $(\BaseFeat)$ & $k$-core & 0.99 & 0.13 \\
\texttt{\RandomVsPrice-2} & $(\BaseFeat)$ & total-degree & 0.11 & -0.79 \\
\texttt{\RandomVsPrice-3} & $(\BaseFeat)$ & in-degree & -0.06 & 0.43 \\
\noalign{\vskip 4pt} 
\multicolumn{5}{l}{
\bgcolored{
$\GraphT$: p2p-Gnutella08 , $\GraphB$: Price~2 (Case Study 1)}{0.62\linewidth}
} \\ 
\texttt{\PtoPVsPriceTwo-1} & $(\BaseFeat)$ & $k$-core &  0.97 & 0.25 \\
\texttt{\PtoPVsPriceTwo-2} & $(\BaseFeat)$ & total-degree & 0.20 & -0.79 \\
\texttt{\PtoPVsPriceTwo-3} & $(\BaseFeat)$ & in-degree & -0.12 & 0.43 \\
\noalign{\vskip 4pt} 
\multicolumn{5}{l}{
\bgcolored{
$\GraphT$: p2p-Gnutella08, $\GraphB$: Enhanced Price (Case Study 1)}{0.62\linewidth}
} \\ 
\texttt{\PtoPVsEPrice-1} & $(\BaseFeat)$ & total-degree & -0.82 & 0.03 \\
\texttt{\PtoPVsEPrice-2} & $(\BaseFeat)$ & in-degree & 0.43 & 0.67 \\
\texttt{\PtoPVsEPrice-3} & $(\BaseFeat)$ & Katz & 0.35 & -0.74 \\
\noalign{\vskip 4pt} 
\multicolumn{5}{l}{
\bgcolored{
$\GraphT$: LC-multiple , $\GraphB$: Combined-AP/MS (Case Study 2)}{0.62\linewidth}
} \\ 
\texttt{\LCMultiVsCombinedAPMS-1} & $(\RelFeatOpe{}{\rm \Mean})(\BaseFeat)$ & Katz & 0.83 & 0.01 \\
\texttt{\LCMultiVsCombinedAPMS-2} & $(\RelFeatOpe{}{\rm \Mean})(\BaseFeat)$ & eigenvector & -0.44 & -0.03 \\
\texttt{\LCMultiVsCombinedAPMS-3} & $(\RelFeatOpe{}{\rm \Mean})(\BaseFeat)$ & total-degree & -0.33 & 0.06 \\
\noalign{\vskip 4pt} 
\multicolumn{5}{l}{
\bgcolored{
$\GraphT$: School-Day2 , $\GraphB$: School-Day1 (Case Study 3)}{0.62\linewidth}
} \\ 
\texttt{\SchSecondVsFirst-1} & $(\RelFeatOpe{}{\rm \Mean} \circ \RelFeatOpe{}{\rm \Max})(\BaseFeat)$ & PageRank &  -0.53 &  -0.21 \\
\texttt{\SchSecondVsFirst-2} & $(\RelFeatOpe{}{\rm \Mean} \circ \RelFeatOpe{}{\rm \Max})(\BaseFeat)$ & closeness & -0.40 & 0.33 \\
\texttt{\SchSecondVsFirst-3} & $(\RelFeatOpe{}{\rm \Mean} \circ \RelFeatOpe{}{\rm \Max})(\BaseFeat)$ & betweenness & 0.33 & 0.09 \\

\bottomrule
\end{tabular}

%% file: tables/top_loadings_pc2.tex
\begin{tabular}{rllrr}
\toprule
ID & relational function $\RelFunc$ & base feature $\BaseFeat$ & \texttt{cPC 1} & \texttt{cPC 2} \\
\midrule

\texttt{\LCMultiVsCombinedAPMS-4} & $(\BaseFeat)$ & Katz & 0.00 & -0.81 \\
\texttt{\LCMultiVsCombinedAPMS-5} & $(\BaseFeat)$ & total-degree & -0.05 & 0.42 \\
\texttt{\LCMultiVsCombinedAPMS-6} & $(\BaseFeat)$ & eigenvector & 0.03 & 0.41 \\

\bottomrule
\end{tabular}

%% file: 5_2_design_comparison.tex
\clearpage

\subsection{Comparison with Other Potential Designs}
\label{sec:design_comparisons}

Our i-cNRL utilizes DeepGL and cPCA for cNRL's two essential components---NRL and contrastive learning---to provide interpretable results.
However, if the interpretability is not required, we can replace each of the learning methods with other alternatives. 
Here we compare three different designs for cNRL: (1) DeepGL \& cPCA, (2) GraphSAGE~\citep{hamilton2017inductive} \& cPCA, and (3) DeepGL \& cVAE~\citep{abid2019contrastive}. 

\subsubsection{Quantitative Results}

We compare the quality of contrastive representations obtained with each design. 
A good contrastive representation should more widely distribute nodes in the target network than the background, and it should also show different patterns in the target and background networks. 
For example, as shown in \autoref{fig:multiple_alphas}, cPCA ($\ContParam=72$) provides a better contrastive representation than PCA ($\ContParam=0$).
To compare the aspects above, we use three different dissimilarity measures: dispersion ratio, Bhattacharyya distance~\citep{bi2017uncertainty}, and Kullback-Leibler (KL) divergence~\citep{wang2009divergence} from a set of nodes in $\ContReprB$ to that in $\ContReprT$. 
The dispersion ratio represents how widely nodes in $\ContReprT$ are scattered relative to $\ContReprB$. 
The Bhattacharyya distance indicates closeness or overlaps of nodes in $\ContReprT$ and $\ContReprB$. 
The KL divergence of $\ContReprT$ from $\ContReprB$ shows the difference between their probability distributions of nodes.
For all the above measures, the higher the value, the better the design.

We calculate the dispersion ratio of $\ContReprT$ to $\ContReprB$ with:
$\frac{\tr(\ContReprT'^\top \ContReprT')/ \nNodesT}{  \tr(\ContReprB'^\top \ContReprB') / \nNodesB}$,
where $\ContReprT'$ and $\ContReprB'$ are the scaled matrices of $\ContReprT$ and $\ContReprB$, respectively, obtained by applying the standardization to a concatenated matrix of $\ContReprT$ and $\ContReprB$.
We use $\ContReprT'$ and $\ContReprB'$, instead of  $\ContReprT$ and $\ContReprB$, to avoid the scaling differences in the embedding's axes across the three designs.
For the Bhattacharyya distance and KL divergence, since we do not have the exact probability distributions of $\ContReprT$ and $\ContReprB$, we employ the estimation methods described by \citet{bi2017uncertainty} and \citet{wang2009divergence}.

For GraphSAGE, we specifically select the GraphSAGE-maxpool model because it produces the best result according to \citet{hamilton2017inductive}.
We use the default parameter values used by \citet{hamilton2017inductive} and \citet{abid2019contrastive} for GraphSAGE and cVAE, except that we set $24$ as the number of features leaned by GraphSAGE. 
For the input features of GraphSAGE, we set the same base features used for DeepGL (see \autoref{table:cnrl_detail_settings} for details).
We obtain 2D embeddings with the cPCs (with cPCA) or \textit{salient latent variables} (with cVAE)~\citep{abid2019contrastive}. 
Since cVAE relies on the probabilistic encoders, the results could be different for each trial, and thus we compute the mean value of each measure for 10 trials.

\begin{table}[tb]
\renewcommand{\tabcolsep}{5pt}
\footnotesize
\centering
\caption{Comparison of contrastive representation quality.}
\label{table:design_comparison}
\input{tables/design_comparison.tex}
\end{table}

\autoref{table:design_comparison} shows a comparison of the three methods on different pairs of networks using the measures above. 
We can see that in general DeepGL~\&~cPCA and GraphSAGE~\&~cPCA have better scores than DeepGL~\&~cVAE. 
Between DeepGL~\&~cPCA and GraphSAGE~\&~cPCA, DeepGL~\&~cPCA tends to provide better results except for the dolphin and Karate networks, which have small numbers of nodes.  

\begin{figure}[tb]
	\centering
    \includegraphics[width=\linewidth]{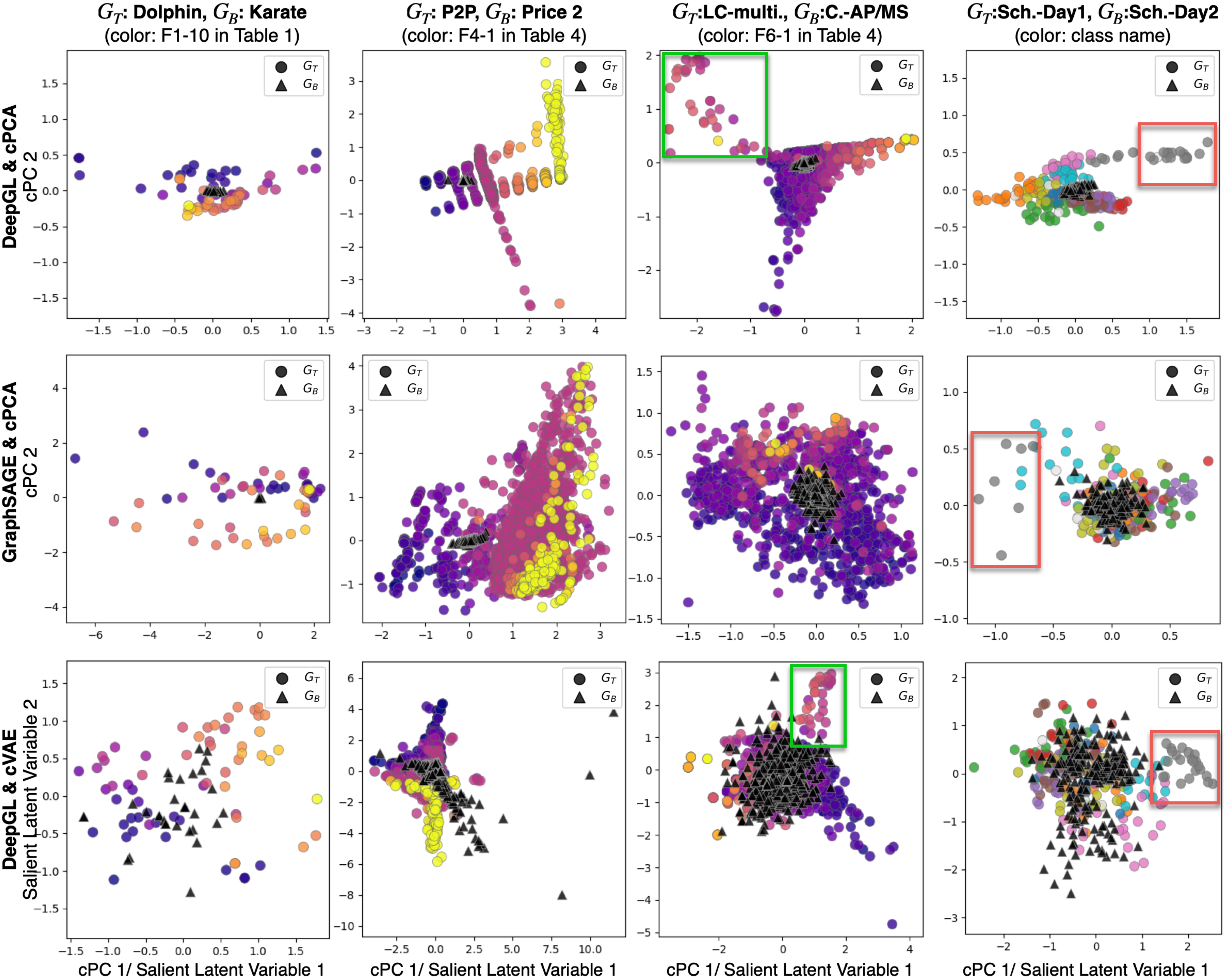}
    \caption{Visual comparison of the 2D embeddings. In the left three columns, the nodes in $\GraphT$ are colored based on values of the feature that most contributes to \texttt{cPC~1}. The nodes in the far right column are colored by their class name. 
    The green and red rectangles annotate distinct groups that can be seen in the 2D embeddings (refer to the description in Qualitative Results).
    }
	\label{fig:design_comparison}
\end{figure}

\subsubsection{Qualitative Results}

We visually compare the embedding results to review more detailed differences, as shown in \autoref{fig:design_comparison}. 
For cVAE, we show the results that have the longest Bhattacharyya distance from 10 trials.
Because GraphSAGE and cVAE do not provide interpretable features, for the comparison, we colorcode the nodes of the target network by the feature values from the DeepGL results.
In specific, the left three columns in \autoref{fig:design_comparison} are colored based on values of the feature that has the top absolute loadings for \texttt{cPC~1} and the far right column is colored by their class name. 

We can see that although the quality of the contrastive representation in \autoref{table:design_comparison} is different, these different designs seem to identify similar unique patterns. 
For instance, all the results of P2P and Price 2 show monotonic increase of the feature value (\texttt{F4-1}---$k$-core numbers). 
Also, for LC-mupltiple and Combined-AP/MS, both DeepGL~\&~cPCA and DeepGL~\&~cVAE depict clearly separated patterns, as indicated with the green rectangles while GraphSAGE~\&~cPCA does not show the same pattern.
Furthermore, in each result of the school networks, we can see a distinct group that consists of gray nodes, as annotated with the red rectangles.

\sloppy{From the above quantitative and qualitative comparisons, we can see that DeepGL~\&~cPCA (i.e., i-cNRL) generates similar quality results when compared with the alternatives.}
However, the other two designs do not provide interpretable results.

%% file: tables/design_comparison.tex
\begin{tabular}{llrrrrrrrrr}
\toprule
& & \multicolumn{3}{c}{dispersion ratio} & \multicolumn{3}{c}{Bhattacharyya} & \multicolumn{3}{c}{KL of $\ContReprT$ from $\ContReprB$} \\
\cmidrule(lr){3-5} \cmidrule(lr){6-8} \cmidrule(lr){9-11}
& & DG\& & GS\& & DG\&  & DG\& & GS\& & DG\&  & DG\& & GS\& & DG\& \\
$\GraphT$ & $\GraphB$ & cPCA & cPCA & cVAE  & cPCA & cPCA & cVAE & cPCA & cPCA & cVAE \\
\midrule
Dolphin & Karate    & 174 & \textbf{9,754} & 1.48 & 1.40 & \textbf{1.73} & 0.92 & 6.82 & \textbf{12.76} & 0.96 \\
P2P  & Price 2  & \textbf{21,744} & 1,801 & 3.36 & \textbf{7.52} & 4.72 & 1.13 & \textbf{45.73} & 14.09 & 36.4 \\
LC-multi. & C.-AP/MS  & \textbf{376} & 54 & 2.71 & 1.52 & \textbf{1.76} & 0.31 & \textbf{18.49} & 16.61 & 15.09 \\
Sch.-Day2 & Sch.-Day1 & \textbf{57} & 6 & 2.20 & \textbf{1.81} & 0.60 & 0.56 & \textbf{5.82}  &  1.80 & 0.80 \\
\bottomrule
\noalign{\vskip 2pt} 
\multicolumn{11}{l}{*DG=DeepGL, GS=GraphSAGE, P2P=p2p-Gnutella08, C.-AP/MS=Combined-AP/MS}
\end{tabular}

%% file: 8_conclusions.tex
\section{Conclusion and Future Work}

This work introduces contrastive network representation learning (cNRL), which aims to reveal unique patterns in one network relative to another. 
Furthermore, we demonstrate a method of cNRL, i-cNRL, that is generic and interpretable. 
With these contributions, our work provides a new approach to network comparison. 

We have demonstrated the usability of i-cNRL with small- or medium-scale networks (less than 10,000 nodes) to provide intelligible examples. 
As a next step, we plan to apply i-cNRL on larger networks (e.g., networks with millions of nodes). 
When analyzing such large, complex networks, the linearity of cPCA used in i-cNRL might limit the capability of finding unique patterns.
Therefore, we will investigate how to incorporate nonlinear contrastive learning methods (such as cVAE) for cNRL while retaining interpretability.

%% file: A_appendix.tex
\section{Implementation Details}
\label{app:implementation}

We have implemented the cNRL architecture with \proglang{Python 3} (refer to \url{https://github.com/takanori-fujiwara/cnrl}).
The implemented cNRL architecture allows the user to apply any NRL and contrastive learning methods that provide ``\code{fit}'' and ``\code{transform}'' methods (as similar to machine learning methods supported in \pkg{scikit-learn}\footnote{\pkg{scikit-learn}, \url{https://scikit-learn.org/}, accessed 2021-8-11.}).
For the implementation of i-cNRL, we have integrated DeepGL and cPCA into the cNRL architecture. 
Because there is no implementation of DeepGL available from \proglang{Python}\footnote{Implementation using \proglang{Java} with \pkg{Neo4j} database is available from \url{https://github.com/neo4j-graph-analytics/ml-models}, accessed 2021-8-11.}, we have implemented DeepGL with \pkg{graph-tool}\footnote{\pkg{graph-tool}, \url{https://graph-tool.skewed.de/}, accessed 2021-8-11.}.
For cPCA, we have modified the implementation available online\footnote{\pkg{ccpca}, \url{https://github.com/takanori-fujiwara/ccpca}, accessed 2021-8-11.} to add the automatic contrastive automatic selection described in \autoref{sec:icnrl_cl}. 

\section{Datasets}
\label{app:datasets}
For the evaluation, we use the datasets in various data repositories, including SNAP, CCSB Interactome Database, and SocioPatterns as well as the synthetic datasets that we generated. 
To allow the reproducibility of this work, we provide links to the original network datasets, processed datasets, and feature matrices learned by DeepGL and GraphSAGE in \SuppURL{}.

\section{Experiment Details}
\label{app:experiment_details}

The source code for generating the experimental results is available in \SuppURL{}.

\subsection{Learning Parameters of i-cNRL}
\label{app:learning_parameters}

\subsubsection{DeepGL Settings}
DeepGL has multiple adjustable settings as it is introduced as a comprehensive inductive NRL framework. 
We follow the terminologies in the work by \citet{rossi2018deep} to describe the detailed settings for each evaluation.
Refer to the work by \citet{rossi2018deep} for those not explained in this paper (indicated with \textit{italic} fonts below).
For all the cNRL we performed, we have used DeepGL with $h=3$ and the logarithmic binning to feature values with $0.5$ as the \textit{transformation parameter}, but without the \textit{feature diffusion}.
For the other settings, generally, we have used as many different relational feature operators and base features as possible for each network dataset.
As for the relational feature operators, for directed networks, we have used all the combinations of $\{\RelFeatOpe{-}{\SummaryMeasure}, \RelFeatOpe{+}{\SummaryMeasure}, \RelFeatOpe{}{\SummaryMeasure}\}$ with $\SummaryMeasure = \{\Mean, \Sum, \Max, \Lpnorm\}$ (i.e., 12 operators in total). 
For undirected networks, we have used $\RelFeatOpe{}{\SummaryMeasure}$ where $\SummaryMeasure = \{\Mean, \Sum, \Max, \Lpnorm\}$. 
As for the base feature $\BaseFeat$, we have used all centralities and measures available in \pkg{graph-tool}. 
However, for each network, some of these features have produced `\code{NaN}' values (e.g., for a disconnected network, the closeness centrality of each node is `\code{NaN}' because each node does not have a path to some other node).
In that case, we have excluded such features from the base features.
Note that, consequently, i-cNRL would not capture the unique patterns if the patterns are highly related to the excluded features.
This is a limitation of our implementation where the computation of the base feature replies on \pkg{graph-tool}. 
However, if needed, without using our implementation for the base feature computation, analysts can precompute a variant of the centrality that does not produce `\code{NaN}' (e.g., instead of ordinary closeness, using the adjusted closeness by \cite{beauchamp1965improved}).
\autoref{table:cnrl_detail_settings} shows the base features we used for each analysis.
For feature pruning of the learned $\F_i$, we have applied the same method used in the work by \citet{rossi2018deep} with the \textit{feature similarity threshold}, $\lambda$.
As $\lambda$ becomes larger, the number of features learned by NRL (i.e., $d$) increases. 
We have set a different $\lambda$ value for each analysis, as listed in \autoref{table:cnrl_detail_settings}. 
In general, for the undirected networks, we have used relatively higher values ($\lambda=0.7$) because the number of base features used is smaller when compared with the directed networks. 

\begin{table}[tb]
\renewcommand{\tabcolsep}{4pt}
\renewcommand{\arraystretch}{0.9}
\footnotesize
\centering
\caption{The detailed DeepGL settings for each analysis.}
\label{table:cnrl_detail_settings}
\input{tables/cnrl_detail_settings.tex}
\end{table}

\subsubsection{cPCA Settings}
For all results, we have used cPCA with the automatic contrastive parameter selection and default settings. 
That is, we have applied the standardization to each of $\FeatMatT$ and $\FeatMatB$ for both learning and projection and the automatic contrastive parameter selection with $\ContConst=10^{-3}$.

\subsection{Full Sets of cPC Loadings}
\label{app:cpc_loadings}
The full sets of cPC loadings obtained with i-cNRL for each analysis in \autoref{sec:test_models} and \autoref{sec:case_studies} are listed in \autoref{table:all_loadings0}-\ref{table:all_loadings3}.

\begin{table}[tb]
\renewcommand{\arraystretch}{0.8}
\footnotesize
\centering
\caption{All cPC loadings for \autoref{sec:test_models}.}
\label{table:all_loadings0}
\input{tables/all_loadings0.tex}
\end{table}

\begin{table}[tb]
\renewcommand{\arraystretch}{0.8}
\footnotesize
\centering
\caption{All cPC loadings for Case Study 1.}
\label{table:all_loadings1}
\input{tables/all_loadings1.tex}
\end{table}

\begin{table}[tb]
\renewcommand{\arraystretch}{0.8}
\footnotesize
\centering
\caption{All cPC loadings for Case Study 2.}
\label{table:all_loadings2}
\input{tables/all_loadings2.tex}
\end{table}

\begin{table}[tb]
\renewcommand{\arraystretch}{0.8}
\footnotesize
\centering
\caption{All cPC loadings for Case Study 3.}
\label{table:all_loadings3}
\input{tables/all_loadings3.tex}
\end{table}

\subsection{Network Generation Models and Parameters}
\label{sec:net_gen_model_prameters}
We have used the Gilbert's and Price's network models to generate Random (\texttt{\RandomNwID}), Price (\texttt{\PriceNwID}), and Price~2 (\texttt{\PriceTwoNwID}) in \autoref{table:network_info}.
Also, in Case Study 1, we have introduced the enhanced Price's network model as the solution to generate a network of which nodes have different $k$-core numbers---Enhanced Price (\texttt{\EPriceNwID}).
In the following, we explain the details of the parameters we used for the network generation and the enhanced Price's model.

\subsubsection{Parameters for the Gilbert's and Price's Models}
The Gilbert's model generating a random network requires the fixed probability of a connection of each pair of nodes. 
We have set the probability to $0.05$ for generating Random (\texttt{\RandomNwID}).
The Price's model requires the fixed number of out-degree of newly added nodes as its parameter.
We have set this parameter to 3 for both Price (\texttt{\PriceNwID}) and Price 2 (\texttt{\PriceTwoNwID}).

\subsubsection{Enhanced Price's Model}
\label{sec:enhanced_price}
For the enhanced Price's model, we modify the Price's model to be able to generate nodes with various $k$-core numbers. 
To achieve this, in the enhanced Price's model, we allow the user to set multiple positive integer numbers of out-degree of newly added nodes. 
We denote this input as $\kappa=\{\kappa_1, \cdots, \kappa_u\}$ where $u$ is the length of the input.
To select one number from $\kappa$ when a new node is added, we need to set the probability of selecting each number.
We denote the probabilities as $p=\{p_1, \cdots, p_u\}$ where $\sum p = 1$. 

To generate Enhanced Price (\texttt{\EPriceNwID}), we have set these parameters to $\kappa=$\{1, 2, 3, 4, 5, 6, 7, 8, 9, 10\} and $p=$\{0.3, 0.25, 0.15, 0.1, 0.075, 0.05, 0.025, 0.025, 0.0125, 0.0125\}.

\subsection{Settings of GraphSAGE and cVAE}
\label{sec:detail_graphsage_cvae}
We describe the detailed settings and parameters of GraphSAGE and cVAE used in \autoref{sec:design_comparisons}.
We have used the source code provided by the authors of GraphSAGE\footnote{GraphSAGE: \url{https://github.com/williamleif/GraphSAGE}, accessed 2021-8-11.} and cVAE\footnote{Contrastive VAE: \url{https://github.com/abidlabs/contrastive_vae}, accessed 2021-8-11.}. 
For GraphSAGE, we have used the unsupervised model \code{graphsage\_maxpool} with 24 as the number of features learned (i.e., \code{dim\_1 = 12} and \code{dim\_2 = 12}) while we have followed the default values for other parameters (e.g., \code{learning\_rate = 0.00001} and \code{model\_size = `small'}). 
We have used cVAE with the default parameters (i.e., \code{intermediate\_dim = 12}, \code{latent\_dim =     2}, \code{batch\_size = 64}, and \code{epochs = 500}).

\subsection{Automatic Contrastive Parameter Selection}
\label{app:auto_alpha}

\begin{figure}[ht]
	\centering
	\captionsetup{farskip=0pt}
    \includegraphics[width=\linewidth]{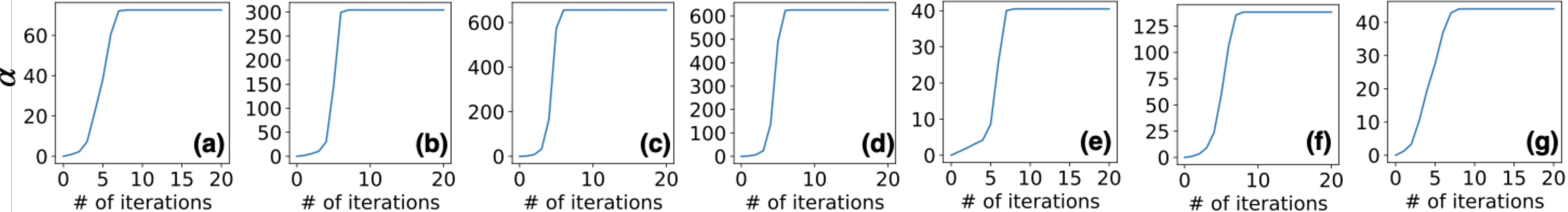}
    \caption{Transitions of $\ContParam$ with the automatic selection: (a) $\GraphT$: Dolphin, $\GraphB$: Karate, (b) $\GraphT$: Price, $\GraphB$: Random, (c) $\GraphT$: Random, $\GraphB$: Price, (d) $\GraphT$: p2p-Gnutella08, $\GraphB$: Price 2, (e) $\GraphT$: p2p-Gnutella08, $\GraphB$: Enhanced Price, (f) $\GraphT$: LC-multiple, $\GraphB$: Combined-AP/MS,
    and (g) $\GraphT$: School-Day2, $\GraphB$: School-Day2.}
	\label{fig:auto_alpha}
\end{figure}

\autoref{fig:auto_alpha} shows transitions of $\ContParam$ value during the automatic selection in i-cNRL. 
For all the experiments, we can see that $\ContParam$ reaches the convergence before 10 iterations.

%% file: tables/cnrl_detail_settings.tex
\definecolor{bgcolor}{rgb}{0.9,0.9,0.9}

\begin{tabular}{p{0.16\linewidth}p{0.19\linewidth}p{0.5\linewidth}p{0.03\linewidth}}
\toprule
$\GraphT$ & $\GraphB$ & $\BaseFeat$ & $\lambda$\\
\hline
Dolphin & Karate & \{total-degree, betweenness, closeness, eigenvector, PageRank, Katz\} & 0.7 \\
\noalign{\vskip 5pt} 
Price & Random & \{in-degree, out-degree, total-degree, PageRank, \newline betweenness, Katz, $k$-core\} & 0.3 \\
\noalign{\vskip 5pt} 
Random & Price & \{in-degree, out-degree, total-degree, PageRank,  \newline betweenness, Katz, $k$-core\} & 0.3 \\
\noalign{\vskip 5pt} 
p2p-Gnutella08 & Price 2 & \{in-degree, out-degree, total-degree, PageRank, \newline betweenness, Katz, $k$-core\} & 0.5 \\
\noalign{\vskip 5pt} 
p2p-Gnutella08 & Enhanced Price & \{in-degree, out-degree, total-degree, PageRank, \newline betweenness, Katz, $k$-core\} & 0.5 \\
\noalign{\vskip 5pt} 
LC-multiple & Combined-AP/MS & \{total-degree, betweenness, eigenvector,  PageRank, Katz\} & 0.7 \\
\noalign{\vskip 5pt} 
School-Day2 & School-Day1 & \{gender, total-degree, closeness, betweenness,   eigenvector, PageRank, Katz\} & 0.7 \\
\bottomrule
\end{tabular}

%% file: tables/all_loadings0.tex
\definecolor{bgcolor}{rgb}{0.9,0.9,0.9}

\begin{tabular}{llrr}
\toprule
relational function $\RelFunc$ & base feature $\BaseFeat$ & \texttt{cPC 1} & \texttt{cPC 2} \\
\midrule

\multicolumn{4}{l}{
\bgcolored{$\GraphT$: Price, $\GraphB$: Random}{0.54\linewidth}
} \\
\input{./tables/loading_data/price_rand_loadings.tex}

\multicolumn{4}{l}{
\bgcolored{$\GraphT$: Random, $\GraphB$: Price}{0.54\linewidth}
} \\ 
\input{./tables/loading_data/rand_price_loadings.tex}

\bottomrule
\end{tabular}

%% file: tables/loading_data/price_rand_loadings.tex
                                                           $(\BaseFeat)$ &                in-degree &   0.23 &   0.71 \\
                                                           $(\BaseFeat)$ &               out-degree &   0.52 &   0.00 \\
                                                           $(\BaseFeat)$ &             total-degree &  -0.79 &  -0.01 \\
                                                           $(\BaseFeat)$ &                 PageRank &   0.00 &   0.00 \\
                                                           $(\BaseFeat)$ &              betweenness &   0.00 &   0.00 \\
                                                           $(\BaseFeat)$ &                     Katz &   0.24 &  -0.71 \\
                                                           $(\BaseFeat)$ &      $k$-core &   0.00 &   0.00 \\
                                $(\RelFeatOpe{-}{\rm \Mean})(\BaseFeat)$ &                in-degree &  -0.01 &   0.01 \\
$(\RelFeatOpe{-}{\rm \Mean} \circ \RelFeatOpe{-}{\rm \Mean})(\BaseFeat)$ &                in-degree &  -0.00 &   0.00 \\

%% file: tables/loading_data/rand_price_loadings.tex
                                                           $(\BaseFeat)$ &                in-degree &  -0.06 &   0.43 \\
                                                           $(\BaseFeat)$ &               out-degree &   0.02 &   0.04 \\
                                                           $(\BaseFeat)$ &             total-degree &   0.11 &  -0.79 \\
                                                           $(\BaseFeat)$ &                 PageRank &   0.02 &  -0.01 \\
                                                           $(\BaseFeat)$ &              betweenness &  -0.01 &   0.00 \\
                                                           $(\BaseFeat)$ &                     Katz &  -0.05 &   0.40 \\
                                                           $(\BaseFeat)$ &      $k$-core &   0.99 &   0.13 \\
                                $(\RelFeatOpe{-}{\rm \Mean})(\BaseFeat)$ &                in-degree &  -0.00 &  -0.00 \\
$(\RelFeatOpe{-}{\rm \Mean} \circ \RelFeatOpe{-}{\rm \Mean})(\BaseFeat)$ &                in-degree &  -0.00 &  -0.00 \\

%% file: tables/all_loadings1.tex
\definecolor{bgcolor}{rgb}{0.9,0.9,0.9}

\begin{tabular}{llrr}
\toprule
relational function $\RelFunc$ & base feature $\BaseFeat$ & \texttt{cPC 1} & \texttt{cPC 2} \\
\midrule

\multicolumn{4}{l}{
\bgcolored{
$\GraphT$: p2p-Gnutella08 , $\GraphB$: Price~2}{0.54\linewidth}
} \\ 
\input{./tables/loading_data/p2p_price_loadings.tex}

\multicolumn{4}{l}{
\bgcolored{
$\GraphT$: p2p-Gnutella08 , $\GraphB$: Enhanced Price}{0.54\linewidth}
} \\ 
\input{./tables/loading_data/p2p_price2_loadings.tex}

\bottomrule
\end{tabular}

%% file: tables/loading_data/p2p_price_loadings.tex
                                                           $(\BaseFeat)$ &                in-degree &  -0.12 &   0.43 \\
                                                           $(\BaseFeat)$ &               out-degree &   0.04 &   0.01 \\
                                                           $(\BaseFeat)$ &             total-degree &   0.20 &  -0.79 \\
                                                           $(\BaseFeat)$ &                 PageRank &   0.05 &  -0.00 \\
                                                           $(\BaseFeat)$ &              betweenness &  -0.00 &   0.00 \\
                                                           $(\BaseFeat)$ &                     Katz &  -0.09 &   0.37 \\
                                                           $(\BaseFeat)$ &      $k$-core &   0.97 &   0.25 \\
                                $(\RelFeatOpe{-}{\rm \Mean})(\BaseFeat)$ &                in-degree &  -0.00 &  -0.00 \\
                                $(\RelFeatOpe{-}{\rm \Mean})(\BaseFeat)$ &               out-degree &   0.00 &  -0.00 \\
                                $(\RelFeatOpe{-}{\rm \Mean})(\BaseFeat)$ &              betweenness &   0.00 &  -0.00 \\
                                 $(\RelFeatOpe{}{\rm \Mean})(\BaseFeat)$ &               out-degree &  -0.00 &   0.00 \\
$(\RelFeatOpe{-}{\rm \Mean} \circ \RelFeatOpe{-}{\rm \Mean})(\BaseFeat)$ &                in-degree &  -0.00 &   0.00 \\
$(\RelFeatOpe{-}{\rm \Mean} \circ \RelFeatOpe{-}{\rm \Mean})(\BaseFeat)$ &               out-degree &   0.00 &   0.00 \\
 $(\RelFeatOpe{}{\rm \Mean} \circ \RelFeatOpe{-}{\rm \Mean})(\BaseFeat)$ &               out-degree &   0.00 &  -0.00 \\

%% file: tables/loading_data/p2p_price2_loadings.tex
                                                           $(\BaseFeat)$ &                in-degree &   0.43 &   0.67 \\
                                                           $(\BaseFeat)$ &               out-degree &   0.16 &  -0.01 \\
                                                           $(\BaseFeat)$ &             total-degree &  -0.82 &   0.03 \\
                                                           $(\BaseFeat)$ &                 PageRank &  -0.02 &   0.04 \\
                                                           $(\BaseFeat)$ &              betweenness &   0.00 &  -0.00 \\
                                                           $(\BaseFeat)$ &                     Katz &   0.35 &  -0.74 \\
                                                           $(\BaseFeat)$ &      $k$-core &   0.01 &  -0.01 \\
                                $(\RelFeatOpe{-}{\rm \Mean})(\BaseFeat)$ &                in-degree &  -0.01 &   0.01 \\
                                $(\RelFeatOpe{-}{\rm \Mean})(\BaseFeat)$ &               out-degree &   0.00 &   0.00 \\
                                $(\RelFeatOpe{-}{\rm \Mean})(\BaseFeat)$ &              betweenness &  -0.00 &  -0.00 \\
                                 $(\RelFeatOpe{}{\rm \Mean})(\BaseFeat)$ &               out-degree &   0.00 &   0.00 \\
$(\RelFeatOpe{-}{\rm \Mean} \circ \RelFeatOpe{-}{\rm \Mean})(\BaseFeat)$ &                in-degree &   0.00 &  -0.00 \\
$(\RelFeatOpe{-}{\rm \Mean} \circ \RelFeatOpe{-}{\rm \Mean})(\BaseFeat)$ &               out-degree &   0.00 &  -0.00 \\
 $(\RelFeatOpe{}{\rm \Mean} \circ \RelFeatOpe{-}{\rm \Mean})(\BaseFeat)$ &               out-degree &   0.00 &   0.00 \\

%% file: tables/all_loadings2.tex
\definecolor{bgcolor}{rgb}{0.9,0.9,0.9}

\begin{tabular}{llrr}
\toprule
relational function $\RelFunc$ & base feature $\BaseFeat$ & \texttt{cPC 1} & \texttt{cPC 2} \\
\midrule

\input{./tables/loading_data/lc_collins_loadings.tex}

\bottomrule
\end{tabular}

%% file: tables/loading_data/lc_collins_loadings.tex
                                                         $(\BaseFeat)$ &             total-degree &  -0.05 &   0.42 \\
                                                         $(\BaseFeat)$ &              betweenness &  -0.00 &  -0.00 \\
                                                         $(\BaseFeat)$ &              eigenvector &   0.03 &   0.41 \\
                                                         $(\BaseFeat)$ &                 PageRank &   0.02 &   0.00 \\
                                                         $(\BaseFeat)$ &                     Katz &   0.00 &  -0.81 \\
                               $(\RelFeatOpe{}{\rm \Mean})(\BaseFeat)$ &             total-degree &  -0.33 &   0.06 \\
                               $(\RelFeatOpe{}{\rm \Mean})(\BaseFeat)$ &              betweenness &   0.00 &   0.00 \\
                               $(\RelFeatOpe{}{\rm \Mean})(\BaseFeat)$ &              eigenvector &  -0.44 &  -0.03 \\
                               $(\RelFeatOpe{}{\rm \Mean})(\BaseFeat)$ &                 PageRank &  -0.01 &  -0.01 \\
                               $(\RelFeatOpe{}{\rm \Mean})(\BaseFeat)$ &                     Katz &   0.83 &   0.01 \\
                                $(\RelFeatOpe{}{\rm \Max})(\BaseFeat)$ &              betweenness &   0.01 &   0.00 \\
                                $(\RelFeatOpe{}{\rm \Max})(\BaseFeat)$ &                 PageRank &  -0.02 &   0.00 \\
$(\RelFeatOpe{}{\rm \Mean} \circ \RelFeatOpe{}{\rm \Mean})(\BaseFeat)$ &             total-degree &  -0.08 &  -0.03 \\
$(\RelFeatOpe{}{\rm \Mean} \circ \RelFeatOpe{}{\rm \Mean})(\BaseFeat)$ &              betweenness &   0.00 &  -0.00 \\
$(\RelFeatOpe{}{\rm \Mean} \circ \RelFeatOpe{}{\rm \Mean})(\BaseFeat)$ &                 PageRank &  -0.02 &  -0.02 \\
 $(\RelFeatOpe{}{\rm \Mean} \circ \RelFeatOpe{}{\rm \Max})(\BaseFeat)$ &              betweenness &   0.00 &   0.00 \\
 $(\RelFeatOpe{}{\rm \Mean} \circ \RelFeatOpe{}{\rm \Max})(\BaseFeat)$ &                 PageRank &   0.03 &   0.03 \\
  $(\RelFeatOpe{}{\rm \Max} \circ \RelFeatOpe{}{\rm \Max})(\BaseFeat)$ &              betweenness &  -0.00 &  -0.01 \\
  $(\RelFeatOpe{}{\rm \Max} \circ \RelFeatOpe{}{\rm \Max})(\BaseFeat)$ &                 PageRank &   0.00 &  -0.01 \\

%% file: tables/all_loadings3.tex
\definecolor{bgcolor}{rgb}{0.9,0.9,0.9}

\begin{tabular}{llrr}
\toprule
relational function $\RelFunc$ & base feature $\BaseFeat$ & \texttt{cPC 1} & \texttt{cPC 2} \\
\midrule

\input{./tables/loading_data/day2_day1_loadings.tex}

\bottomrule
\end{tabular}

%% file: tables/loading_data/day2_day1_loadings.tex
                                                          $(\BaseFeat)$ &             total-degree &  -0.18 &  -0.29 \\
                                                          $(\BaseFeat)$ &                closeness &   0.03 &  -0.00 \\
                                                          $(\BaseFeat)$ &              betweenness &   0.03 &  -0.01 \\
                                                          $(\BaseFeat)$ &              eigenvector &   0.22 &  -0.18 \\
                                                          $(\BaseFeat)$ &                 PageRank &  -0.01 &   0.32 \\
                                                          $(\BaseFeat)$ &                     Katz &  -0.08 &   0.14 \\
                                                          $(\BaseFeat)$ &                   gender &   0.02 &   0.01 \\
                                $(\RelFeatOpe{}{\rm \Mean})(\BaseFeat)$ &             total-degree &  -0.24 &  -0.10 \\
                                $(\RelFeatOpe{}{\rm \Mean})(\BaseFeat)$ &              betweenness &   0.08 &   0.06 \\
                                $(\RelFeatOpe{}{\rm \Mean})(\BaseFeat)$ &                   gender &   0.04 &   0.02 \\
                                 $(\RelFeatOpe{}{\rm \Sum})(\BaseFeat)$ &                   gender &  -0.02 &  -0.01 \\
                                 $(\RelFeatOpe{}{\rm \Max})(\BaseFeat)$ &             total-degree &   0.04 &  -0.23 \\
                                 $(\RelFeatOpe{}{\rm \Max})(\BaseFeat)$ &                closeness &  -0.12 &   0.02 \\
                                 $(\RelFeatOpe{}{\rm \Max})(\BaseFeat)$ &              betweenness &   0.06 &   0.00 \\
                                 $(\RelFeatOpe{}{\rm \Max})(\BaseFeat)$ &              eigenvector &   0.12 &  -0.07 \\
                                 $(\RelFeatOpe{}{\rm \Max})(\BaseFeat)$ &                 PageRank &  -0.10 &   0.10 \\
                                 $(\RelFeatOpe{}{\rm \Max})(\BaseFeat)$ &                     Katz &   0.03 &   0.17 \\
                                 $(\RelFeatOpe{}{\rm \Max})(\BaseFeat)$ &                   gender &   0.00 &   0.00 \\
                              $(\RelFeatOpe{}{\rm \Lpnorm})(\BaseFeat)$ &                   gender &  -0.02 &  -0.01 \\
 $(\RelFeatOpe{}{\rm \Mean} \circ \RelFeatOpe{}{\rm \Mean})(\BaseFeat)$ &             total-degree &   0.23 &   0.09 \\
 $(\RelFeatOpe{}{\rm \Mean} \circ \RelFeatOpe{}{\rm \Mean})(\BaseFeat)$ &              betweenness &  -0.14 &  -0.02 \\
 $(\RelFeatOpe{}{\rm \Mean} \circ \RelFeatOpe{}{\rm \Mean})(\BaseFeat)$ &                   gender &  -0.09 &  -0.05 \\
  $(\RelFeatOpe{}{\rm \Mean} \circ \RelFeatOpe{}{\rm \Max})(\BaseFeat)$ &             total-degree &   0.19 &  -0.38 \\
  $(\RelFeatOpe{}{\rm \Mean} \circ \RelFeatOpe{}{\rm \Max})(\BaseFeat)$ &                closeness &  -0.40 &   0.33 \\
  $(\RelFeatOpe{}{\rm \Mean} \circ \RelFeatOpe{}{\rm \Max})(\BaseFeat)$ &              betweenness &   0.33 &   0.09 \\
  $(\RelFeatOpe{}{\rm \Mean} \circ \RelFeatOpe{}{\rm \Max})(\BaseFeat)$ &              eigenvector &   0.29 &  -0.29 \\
  $(\RelFeatOpe{}{\rm \Mean} \circ \RelFeatOpe{}{\rm \Max})(\BaseFeat)$ &                 PageRank &  -0.53 &  -0.21 \\
  $(\RelFeatOpe{}{\rm \Mean} \circ \RelFeatOpe{}{\rm \Max})(\BaseFeat)$ &                     Katz &   0.22 &   0.50 \\
  $(\RelFeatOpe{}{\rm \Mean} \circ \RelFeatOpe{}{\rm \Max})(\BaseFeat)$ &                   gender &  -0.00 &   0.00 \\
  $(\RelFeatOpe{}{\rm \Max} \circ \RelFeatOpe{}{\rm \Mean})(\BaseFeat)$ &              betweenness &   0.00 &  -0.01 \\
  $(\RelFeatOpe{}{\rm \Max} \circ \RelFeatOpe{}{\rm \Mean})(\BaseFeat)$ &                   gender &  -0.01 &  -0.01 \\
   $(\RelFeatOpe{}{\rm \Max} \circ \RelFeatOpe{}{\rm \Sum})(\BaseFeat)$ &                   gender &  -0.00 &  -0.00 \\
$(\RelFeatOpe{}{\rm \Max} \circ \RelFeatOpe{}{\rm \Lpnorm})(\BaseFeat)$ &                   gender &  -0.00 &  -0.00 \\